\documentclass[preprint,12pt,authoryear]{elsarticle}
\usepackage[margin=1in]{geometry}
\newtheorem{theorem}{Theorem} 
\usepackage[utf8]{inputenc}
\usepackage{amsmath, amssymb}
\usepackage{graphicx}         
\usepackage{float}            
\usepackage{hyperref}         
\usepackage{cleveref}
\usepackage{cite}             
\usepackage{xcolor}
\usepackage{subcaption}
\usepackage{graphicx}
\usepackage{tabularx}
\usepackage{booktabs}
\usepackage{array} 

\date{\today} 

\begin{document}
\begin{frontmatter}

\title{MLPs and KANs for data-driven learning in physical problems: A performance comparison} 

\author[label1]{Raghav Pant}
\ead{raghav.pant@utexas.edu}

\author[label2]{Sikan Li}
\ead{sli@tacc.utexas.edu}

\author[label1]{Xingjian Li}
\ead{xingjian.li@austin.utexas.edu}

\author[label1]{Hassan Iqbal}
\ead{hassan.iqbal@utexas.edu}

\author[label1,label3]{Krishna Kumar}
\ead{krishnak@utexas.edu}

\affiliation[label1]{organization={The Oden Institute for Computational Engineering and Sciences}, addressline={The University of Texas at Austin},country={USA}}
\affiliation[label2]{organization={Texas Advanced Computing Center (TACC)},addressline={The University of Texas at Austin},country={USA}}
\affiliation[label3]{organization={Department of Civil, Architecture and Environmental Engineering}, addressline={The University of Texas at Austin},country={USA}}

\begin{abstract}

There is increasing interest in solving partial differential equations (PDEs) by casting them as machine learning problems. Recently, there has been a spike in exploring Kolmogorov-Arnold Networks (KANs) as an alternative to traditional neural networks represented by Multi-Layer Perceptrons (MLPs). While showing promise, their performance advantages in physics-based problems remain largely unexplored. Several critical questions persist: Can KANs capture complex physical dynamics and under what conditions might they outperform traditional architectures? In this work, we present a comparative study of KANs and MLPs for learning physical systems governed by PDEs. We assess their performance when applied in deep operator networks (DeepONet) and graph network-based simulators (GNS), and test them on physical problems that vary significantly in scale and complexity. Drawing inspiration from the Kolmogorov Representation Theorem, we examine the behavior of KANs and MLPs across shallow and deep network architectures. Our results reveal that although KANs do not consistently outperform MLPs when configured as deep neural networks, they demonstrate superior expressiveness in shallow network settings, significantly outpacing MLPs in accuracy over our test cases. This suggests that KANs are a promising choice, offering a balance of efficiency and accuracy in applications involving physical systems.

\end{abstract}

\end{frontmatter}

\section{Introduction}

Simulating physical systems governed by partial differential equations (PDEs) or complex dynamical laws is fundamental across scientific and engineering disciplines. Traditional numerical methods like finite elements provide reliable solutions, but often incur prohibitive computational costs, especially for high-dimensional problems or scenarios requiring repeated simulations. Neural network-based approaches have emerged as compelling alternatives, offering accelerated simulations by learning surrogate models or solution operators directly from data \citep{karniadakis2021physics}.

The Multi-Layer Perceptron (MLP) has established itself as the dominant neural architecture for these tasks. MLPs consist of layers of affine transformations followed by fixed non-linear activation functions at the nodes. Their theoretical foundation rests on Universal Approximation Theorems (UATs), which establish that sufficiently large MLPs can approximate any continuous function to arbitrary precision \citep{cybenko1989approximation, hornik1991approximation}. This theoretical guarantee, combined with empirical success, has established MLPs as the standard baseline architecture in scientific machine learning.

Recently, \citet{liu2024kankolmogorovarnoldnetworks} revived interest in networks based on the Kolmogorov-Arnold (KA) representation theorem \citep{arnol1957functions, kolmogorov1957representation}. This theorem proves that any continuous multivariate function can be exactly represented through compositions and sums of univariate functions. Kolmogorov-Arnold Networks (KANs) implement this principle using learnable univariate activation functions on network edges, while nodes typically perform summation operations. This architectural distinction from MLPs, which use fixed activations at nodes, suggests KANs more efficiently represent functions decomposable into simpler components -- a property particularly relevant for physical systems governed by well-defined mathematical principles. (For background on the KA theorem and related concepts, see \citep{hecht1987kolmogorov, igelnik2003kolmogorov, coppejans2004kolmogorov}; for theoretical extensions, see \citep{sprecher1965structure, sprecher1972improvement, sprecher1996numerical, sprecher2017algebra}.)

This fundamental architectural difference raises a critical question: \textit{Do KANs provide meaningful performance advantages over MLPs for data-driven learning of physical systems?} While theoretical properties are suggestive, empirical evaluation is essential, as theoretical advantages do not always translate to practical performance gains.
Previous comparative studies offer limited guidance for data-driven applications. Many investigations have focused on Physics-Informed Neural Networks (PINNs), which integrate physical laws into the loss function \citep{shukla2024comprehensivefaircomparisonmlp, guilhoto2024deep}. However, performance in PINN frameworks may not generalize to purely data-driven settings. Other work has suggested that KANs excel in specific tasks like symbolic regression \citep{yu2024kan}, but their advantages in learning physical operators remain contested. \citet{shukla2024comprehensivefaircomparisonmlp} reported limited benefits from KANs in operator learning, while \citet{abueidda2025deepokan} demonstrated KANs outperforming MLPs under specific configurations. This inconsistency underscores the need for systematic comparison within established data-driven frameworks like Deep Operator Networks (DeepONet) \citep{lu2021learning} and Graph Network-based Simulators (GNS) \citep{sanchez2020learning}, with particular attention to network depth as a critical factor.

We hypothesize that the relative performance of KANs and MLPs depends significantly on network depth, a consideration motivated by their different theoretical foundations (KA theorem versus UAT). To test this hypothesis, we structure our investigation around two distinct configurations: 
\begin{enumerate} 
\item \textbf{Shallow Networks:} Comparing single-hidden-layer MLPs against shallow KANs with architectures informed by the KA representation theorem. 
\item \textbf{Deep Networks:} Evaluating standard deep MLPs against deep KANs constructed by stacking multiple KAN layers. 
\end{enumerate}

For our empirical tests, we employ DeepONet for operator learning tasks derived from PDEs and GNS for learning particle dynamics, both representing state-of-the-art data-driven approaches for physical system modeling.

This work makes three main contributions: 
\begin{itemize} 
\item We provide the first comprehensive empirical comparison between shallow KANs and MLPs across diverse physics-based learning tasks, revealing KANs' superior representation capabilities in shallow network configurations. 

\item We analyze the relative performance of deep KANs versus deep MLPs for operator learning, examining how performance advantages evolve with increasing network depth. 

\item We present the first integration and evaluation of KANs within the GNS message-passing framework for learning spatio-temporal dynamics of particle-based systems, demonstrating their effectiveness for large-scale physical simulations. 
\end{itemize}

The remainder of this paper is organized as follows. In \cref{sec:background}, we provide background information on multilayer perceptrons (MLPs), Kolmogorov-Arnold networks (KANs), deep operator networks (DeepONet) and graph network-based simulators (GNS). In \cref{sec:exp_design}, we outline our experimental design, test setup, model configurations, and procedures used in our study. We present numerical results for different test problems in \cref{sec:numericals}. Finally, in \cref{sec:conclusion}, we discuss our findings, summarize the key insights, and suggest potential directions for future research.

\section{Background}
\label{sec:background}
In this section, we provide the necessary mathematical background that motivates and serves as the foundation of this work. We divide it into four parts. First, we introduce two different neural network architectures: the multilayer perceptron (MLP) and the Kolmogorov–Arnold network (KAN), along with the corresponding representation theorems that guide the design of our experiments. 
Next, we provide a brief introduction to deep operator networks (DeepONet) and graph network-based simulators (GNS), both of which are popular data-driven methods widely used in physics-based applications and crucial for our experiments. While both approaches leverage data to model complex physical systems, they differ significantly in their underlying methodologies, problem scales, and system representations. DeepONet is particularly effective for learning operators that map functions to functions, making it well suited for solving partial differential equations (PDEs) under different input conditions; however, the method generally focuses on problems of smaller scale. In contrast, GNS excel at modeling dynamic, multi-body systems by leveraging graph structures to represent local interactions, and are especially advantageous for performing long-time horizon simulations, a key feature for studying evolving physical phenomena. This distinction highlights the complementary nature of DeepONet and GNS, making them valuable benchmarks for evaluating data-driven approaches in diverse physics-based scenarios.
    
\subsection{Multilayer Perceptron (MLP)}
A \textbf{Multilayer Perceptron (MLP)}~\citep{rumelhart1986learning, haykin1994neural}, sometimes also referred to as a feedforward neural network, is a simple yet versatile neural network architecture widely used in various tasks, including classification, regression, and more. An MLP consists of multiple layers of neurons, where each layer applies a learned transformation to its input using weight matrices and biases. 
The output of an MLP typically takes the following form:
\[
MLP(x) = \sigma\left( W_{L} \sigma\left( W_{L-1} \dots \sigma\left( W_{1} x + b_{1} \right) \dots + b_{L-1} \right) + b_{L} \right).
\]
Here, \( MLP(x) \) denotes the output of the neural network, where the input \(x = (x_1, x_2, \dots, x_n) \) represents the input, typically vectorized. \( W_l \in \mathbb{R}^{k \times n} \) and \( b_l \in \mathbb{R}^k \) are the weight and bias matrices for the \( l \)-th layer, which can vary in size. \( L \) represents the total number of layers excluding the input layer, and \( \sigma \) is the activation function, which is crucial as it introduces nonlinearity into the model. For regression tasks, the activation function in the final layer is often omitted to allow for unrestricted output values.

A fundamental theoretical result that underscores the power of MLPs and justifies their use in numerous applications is the \textbf{Universal Approximation Theorem (UAT)}. While several different variations of the theorem has been proposed over the years, we hereby mainly focus on two of them. 
\begin{theorem}~\citep{cybenko1989approximation, hornik1991approximation}
Let \( C(K, \mathbb{R}^m) \) denote the set of continuous functions from a compact subset \( K \subseteq \mathbb{R}^n \) to \( \mathbb{R}^m \). Suppose \( \sigma \in C(\mathbb{R}, \mathbb{R}) \) is a non-polynomial activation function applied elementwise. Then, for every function \( f \in C(K, \mathbb{R}^m) \) and every \( \varepsilon > 0 \), there exist an integer \( k \in \mathbb{N} \), weight matrix \( W \in \mathbb{R}^{k \times n} \), bias vector \( b \in \mathbb{R}^{k} \), and output weight matrix \( C \in \mathbb{R}^{m \times k} \) such that  
\[
\sup_{x \in K} \| f(x) -g(x) \| < \varepsilon,
\]
where the function \( g(x) \) is represented by a single-hidden-layer neural network:
\[
g(x) = C \cdot \sigma(Wx + b).
\]
\label{thm:UAT_shallow}
\end{theorem}
\cref{thm:UAT_shallow} provides theoretical justification for shallow neural networks with unbounded width. However, in practice, it can be difficult to determine the optimal size of the neural network, and an overparameterized model can lead to decreased efficiency. A more commonly referenced version of the UAT is the ``arbitrary depth'' version from \citep{lu2017expressive}, which reads
\begin{theorem}
For any Lebesgue-integrable function \( f: \mathbb{R}^n \to \mathbb{R} \) and any \( \varepsilon > 0 \), there exists a fully connected ReLU network with width at most \( k + 4 \) such that the function \( g(x) \) represented by this network satisfies  
\[
\int_{\mathbb{R}^n} |f(x) - g(x)| dx < \varepsilon.
\]
\label{thm:UAT_deep}
\end{theorem}
In both theorems $g(x)$ are MLPs defined earlier.
While \cref{thm:UAT_deep} is originally exclusive to scalar functions and MLPs using the ReLU~\citep{agarap2018deep} activation function, these conditions can be relaxed and refined in later works such as \citep{hanin2017approximating, johnson2019deep, kidger2020universal, park2020minimum}. Similarly, practical challenges remain in constructing and optimizing deep networks for complicated tasks.

\subsection{Kolmogorov–Arnold Networks (KAN)}
Inspired by the Kolmogorov-Arnold representation theorem \citep{kolmogorov1957representation, braun2009constructive}, which states
\begin{theorem}
Let \( f: [0,1]^n \to \mathbb{R} \) be an arbitrary continuous function. Then, there exist continuous univariate functions \( \Phi_{q} \) and \( \phi_{qp} \), where \( q = 1, \dots, 2n+1 \) and \( p = 1, \dots, n \), such that:

\[
f(x_1, x_2, \dots, x_n) = \sum_{q=1}^{2n+1} \Phi_q \left( \sum_{p=1}^{n} \phi_{qp} (x_p) \right),
\]
\end{theorem}
note here, the functions \( \Phi_q \) and \( \phi_{qp} \) depend only on one variable each and are independent of \( f \). 
Kolmogorov–Arnold Networks (KANs) are a class of neural networks designed to approximate continuous functions using a series of nested, weighted summations and nonlinear activation functions. 
From \citep{liu2024kankolmogorovarnoldnetworks} in standard form the KAN architecture has formulation 
\[
KAN(x) = \sum_{i_{L-1}=1}^{n_{L-1}} \phi_{L-1,i_L,i_{L-1}} \left( \sum_{i_{L-2}=1}^{n_{L-2}} \cdots  \left( \sum_{i_1=1}^{n_1} \phi_{1,i_2,i_1} \left( \sum_{i_0=1}^{n_0} \phi_{0,i_1,i_0} (x_{i_0}) \right) \right) \dots \right).
\]
This formulation represents a composition of multiple KAN layers, where each layer applies learnable univariate activation functions to refine the function approximation.
\( L \) denotes the total number of layers, \( \{n_j\}_{j=0}^L \) are the numbers of neurons in the \( j \)-th layer, and \( \phi_{i,j,k} \) are the univariate activation functions. 
Breaking down multivariable functions into simpler, one-dimensional function compositions is a defining feature of the KAN architecture. This added expressiveness enables KAN models to accurately approximate complex functions while using fewer trainable parameters for certain applications such as genomic tasks~\citep{cherednichenko2024kolmogorov}, computer vision~\citep{cheon2024demonstratingefficacykolmogorovarnoldnetworks} and time series problems~\citep{vacarubio2024kolmogorovarnoldnetworkskanstime, nehma2024leveragingkansenhanceddeep}. 

For a KAN layer, activation function $\phi(x_i)$ is defined as
\[
\phi(x_i) = w_b b(x_i) + w_s \text{spline}(x_i),
\]
with basis function $b(x_i)$ and spline function $\text{spline}(x_i)$ taking the form
\begin{equation}
    b(x_i) = \frac{x_i}{1+e^{-x_i}}, \qquad \qquad \text{spline}(x_i) = \sum_{j} c_j B_j(x_i).
\label{eq:KAN_basis_spline}
\end{equation}
In practice $w_b, w_s,c_j$ are all trainable parameters critical to the model's performance. Grid size and polynomial order for $B_j(x_i)$ are also significant but are typically chosen and tuned by the user as hyperparameters. 

In \citep{liu2024kankolmogorovarnoldnetworks} the authors also propose the approximation theory (KAT) which we omit here. Similar to the UAT, the KAT proves the approximation property of KAN architectures and justifies its use in machine learning applications. We also want to point out that 
while the Kolmogorov-Arnold representation theorem does not match the KAN architecture in an exact sense, it serves as the pillar stone for the models performances and is often adopted to design KAN models in practice.

\subsection{DeepONet}
\label{sec:deeponet}
The deep operator network (DeepONet) framework~\citep{lu2021learning} is a prominent method in operator learning, finding wide applicability in physics-based applications, particularly in partial differential equations (PDEs) related problems.
In its standard form, the DeepONet architecture leverages two neural networks: 
the \textit{branch network} captures the dependence of the operator on the input function, and the \textit{trunk network} provides a coordinate-dependent representation of the output.
These networks are then combined to approximate the target operator.

Given an operator $\mathcal{G}$ that maps a function $u \in \mathcal{U}$ to another function $\mathcal{G}(u) \in \mathcal{V}$, the goal of DeepONet is to learn the mapping $\mathcal{G}$. Let $x$ denote the input coordinates in the branch network and $y$ denote the output coordinates in the trunk network. The structure of DeepONet can be summarized as:

\begin{equation}
    \mathcal{G}(u)(y) \approx \sum_{i=1}^p b_i(u) t_i(y),
\end{equation}
where $\{ b_i(u)\}_{i=1}^{p}$ are outputs of the branch network, which encodes the input functional space $\mathcal{U}$, and $\{ t_i(y)\}_{i=1}^{p}$ are the outputs of the trunk network, which evaluates the output function at specific coordinates $y \in \mathcal{D}$. Here, $\mathcal{D}$ is the domain of interest. $p$ is the number of basis functions used in the representation. One thing to note about the branch network is that it can only process the function $u$ at a finite set of points $\{x_1, x_2, \ldots, x_m\}$. Thus, the input to the branch network is  $[u(x_1), u(x_2), \ldots, u(x_m)]$. In practice both networks are learnable with possible architectures including MLP and KAN.

The training process involves minimizing the loss function:
\[
    \mathcal{L}(\theta_b, \theta_t) = \frac{1}{N} \sum_{j=1}^N \| \mathcal{G}(u_j)(y_j) - \sum_{i=1}^p b_i(u_j) t_i(y_j) \|^2,
\]
where $N$ is the number of training samples, and $\theta_b$ and $\theta_t$ are the trainable parameters of the branch and trunk networks, respectively. DeepONet is a robust and widely adopted method with extensive results in physics-based problems~\citep{cao2024deep, oommen2022learning, goswami2022neural}, making it a suitable candidate for our testing. 

We also note that follow-up works to the vanilla DeepONet architecture have been proposed in works such as \citep{he2024sequential, peyvan2024riemannonets, wei2023super, mandl2024separable}; however, they will not be the focus of this paper.

\subsection{Graph Network-Based Simulators}
\label{sec:GNS_intro}

Graph network-based simulators (GNS) are machine learning methods that leverage graph neural network architectures to model and simulate particle and fluid flows. They learn to represent the state of a physical system as a graph consisting of nodes and edges. Graph-based approaches offer permutation invariance, i.e. GNS outputs remain consistent regardless of the order in which nodes are presented, which is crucial for handling unordered data such as in particle systems. GNS learn local interactions between particles in the physical system via a data-driven loss on observed history. Once trained, GNS have demonstrated remarkable success in learning to simulate complex physical dynamics over long time horizons with high accuracy, stability and computational efficiency \citep{pfaff2020learning, sanchez2020learning, choi2024graph}. 

A key advantage of GNS-based approaches is their strong generalization ability, which stems directly from their graph-based architecture that learns interactions within local neighborhoods. By modeling particle dynamics through pairwise relations, these systems capture both the shared physical interactions among particles within the inner domain and the collision dynamics at boundaries. This localized learning approach enables GNS to accurately predict behaviors in test scenarios that differ significantly from the training data, including environments with unseen obstacles, positions and velocities, or different spatial domain ranges. 

\begin{figure}
    \centering
    \includegraphics[width=\linewidth]{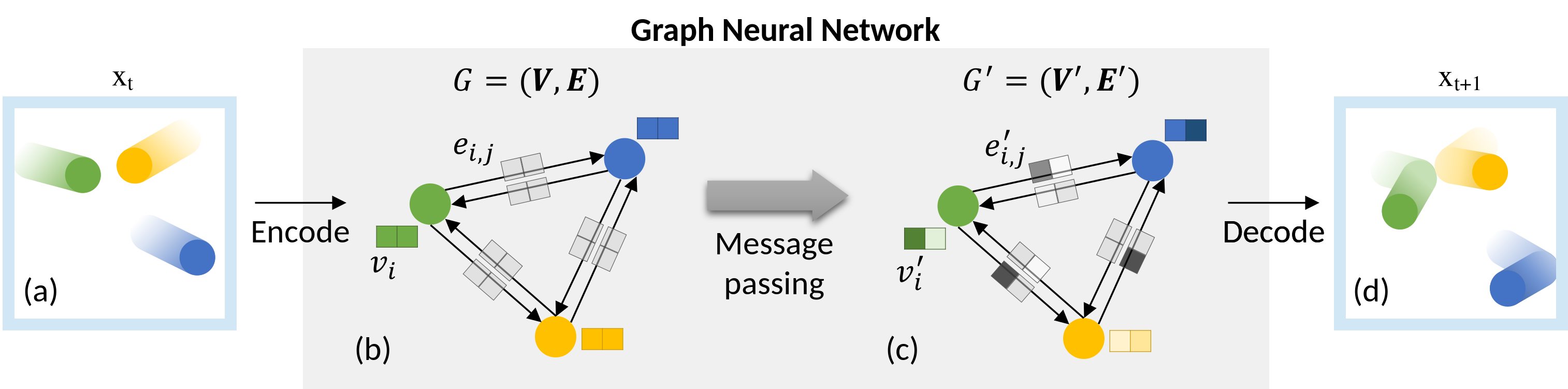}
    \caption{Schematic representation of a GNS model.}
    \label{fig:Gns-schematic}
\end{figure}

The process of graph-based simulation is shown in \cref{fig:Gns-schematic} (modified from \citep{battaglia2018relational}). The current state of physical system $\boldsymbol{x_t}$ is represented as graph  $\Gamma = (\textbf{V},\textbf{E})$, information is propagated through the graph with message passing and the updated graph $\Gamma'=(\textbf{V'},\textbf{E'})$ is decoded to give predicted state of the physical system. For example, in the case of granular soil flows, vertices and edges can represent grains and their directional interactions respectively, and the updated graph outputs the grains positions at the next time step.

Given node feature vectors $\boldsymbol{v}_i$ for all $i \in \mathcal{V}$ and edge feature vectors $\boldsymbol{e}_{i,j}$ for all $ {(i,j)} \in \mathcal{E}$, the graph $\Gamma^{(l)}$ at the $l$-th layer is updated through learned representation given as,

\begin{equation}
    \Gamma'^{(l)} = \underset{\boldsymbol{v}, \boldsymbol{e}}{\mathcal{GNN}}(\Gamma^{(l-1)})
\end{equation}

where $\boldsymbol{v}$ and $\boldsymbol{e}$ are node and edge embeddings respectively. We define the neighborhood indices of node $\boldsymbol{v}_i$ as $\mathcal{N}(i) = \{j \mid (v_j, v_i) \in \mathcal{E} \}$. Hence, the message passing to update node $\boldsymbol{v}_i$ and edge $\boldsymbol{e}_{i,j}$ embeddings can be described as, 

\begin{subequations}
    \begin{align}
        \boldsymbol{v}_i^{(l)} &= \mathbf{f}_{\boldsymbol{\theta}}^{(l)}
        \left(
            \boldsymbol{v}_i^{(l-1)},
            \underset{j \in \mathcal{N}(i)}{\bigoplus} 
            \mathbf{g}_{\boldsymbol{\theta}}^{(l)}
            \left(
                \boldsymbol{v}_j^{(l-1)}, 
                \boldsymbol{v}_i^{(l-1)}, 
                \boldsymbol{e}_{i,j}^{(l-1)}
            \right)
        \right) \\
        \boldsymbol{e}_{i,j}^{(l)}(w, v) &= 
        \mathbf{g}_{\boldsymbol{\theta}}^{(l)}
        \left(
            \boldsymbol{v}_j^{(l-1)}, 
            \boldsymbol{v}_i^{(l-1)}, 
            \boldsymbol{e}_{i,j}^{(l-1)}
        \right).
    \end{align}
    \label{eq:GNS_equations}
\end{subequations}

where $\mathbf{f_\theta}$, $\bigoplus$ and $\mathbf{g_\theta}$ are embedding update function, the aggregate function and message generation function respectively. Here, $\mathbf{f_\theta}$ and $\mathbf{g_\theta}$ are learned functions represented either by MLPs or KANs, while the $\bigoplus$ can be any permutation invariant operator such as summation, mean or maximum \citep{zhao2025physical}.

While PDE operator learning (see \cref{sec:deeponet}) approaches are focused on learning direct maps from initial conditions to solutions at a specified point in time, GNS approach the \textit{simulation} of relatively large-scale particle systems by modeling temporal dynamics through iterative rollouts. GNS typically require larger model architectures to capture the complex spatio-temporal interactions between particles, resulting in significantly higher computational demands than DeepONets.

A given GNS model operates on a high-dimensional latent space determined by the dimensions of node and edge embeddings. While these embeddings capture complex particle properties and interactions, the neural networks used to represent $\mathbf{f_\theta}$ and $\mathbf{g_\theta}$ are typically shallow rather than deep. This architectural choice is deliberate - shallow networks with sufficient width can effectively learn the necessary transformations while maintaining computational efficiency. The key complexity in GNS comes not from network depth but from the iterative message-passing structure and the graph representation itself, which inherently models particle interactions. As the particle count increases, the graph size grows significantly, adding to the computational burden.

There has been some work on leveraging Kolmogorov-Arnold representation in graph neural networks. However, these works are mostly focused on node classification, link prediction, and graph classification \citep{zhang2024graphkan,li2024ka,de2024kolmogorov}. As KANs allow additional expressivity over MLPs in shallow networks, owing to the spline based parameterization, we hypothesize that formulating KANs to model $\mathbf{f_\theta}$ and $\mathbf{g_\theta}$ shall result in accuracy improvements. In this work, we test this hypothesis on two datasets. We present results in \cref{sec:GNS_results}.

\section{Experimental Design}
\label{sec:exp_design}

In these experiments, we aim to compare the performance of MLP and KAN networks across various physics-based applications. Specifically, we focus on solving operator learning problems for differential equations using the DeepONet framework, as well as employing GNS for simulating particle-based dynamical systems. To ensure a comprehensive evaluation, we select examples that encompass problems of varying scale and complexity. Unlike previous studies in this field, we conduct two primary sets of comparisons.

The first comparison involves ``shallow'' networks: a single-layer MLP with a hidden width of $1000$ versus a shallow KAN with one hidden layer of size $2n+1$, where 
$n$ denotes the input dimension of each network (specified for each problem below). This setup is directly inspired by the representation theorem underlying each architecture.  We emphasize the importance of this comparison for two primary reasons. First, KAN architectures demonstrate significantly stronger representation capabilities than MLP models in shallow neural network settings. Second, small and shallow neural networks are frequently employed in physics-based machine learning applications, making this comparison particularly relevant. 

We note that in practice, KANs are often used like MLPs by stacking multiple layers to form ``deep KANs,'' as seen in \citep{liu2024kankolmogorovarnoldnetworks} and subsequent studies \citep{yu2024kan, abueidda2025deepokan}.
To extend our evaluation to common deep networks, we also compare the empirical performance of deep MLPs and deep KANs across all operator learning tasks. Combining the results from both comparisons provides us with unique insights and ensures the validity of our claims. 

While it is common to treat KAN models similarly to MLPs when tuning hyperparameters, we emphasize that adjusting only the depth and width of a model is often insufficient for achieving optimal results. In our work, we expand our hyperparameter choices to include factors such as the spline grid size and the order of polynomials used in defining a KAN model. We observe that increasing the number of spline polynomials can significantly enhance model expressiveness, particularly for shallow neural networks. Consequently, we use $20$ spline polynomials for the trunk networks in our experiments unless stated otherwise, ensuring the best possible results. 

As noted in \cref{sec:GNS_intro}, GNS and similar neural architectures that process information in a high dimensional latent space may benefit from added expressivity of single-layer KANs over shallow MLPs. Therefore, for our study on GNS and particle systems, we solely focus on shallow networks. We construct and test four different GNS models, where the $\mathbf{f_\theta}$ and $\mathbf{g_\theta}$ in Eq. \eqref{eq:GNS_equations}are represented either with a shallow KAN or an MLP. The purpose of these experiments is to evaluate any KAN advantages, both in terms of accuracy and robustness to input noise, within particle dynamics over long time horizon in large-scale systems. We design a suite of experiments across different model configurations and physical problem settings, and employ varying metrics to report insights into large GNS models represented with KANs.

For all of our tested models, we use the Sigmoid Linear Unit (SiLU) activation functions. We avoid layer normalization, dropout, or additional regularization techniques unless specified otherwise to maintain a clear baseline comparison. All experiments are implemented using PyTorch~\citep{paszke2019pytorch}. For the KAN models, we rely extensively on the ``Efficient KAN'' implementation from \citep{efficient_kan}, which enhances the original KAN approach by offering faster computation and reduced memory usage. Model training is performed on a single NVIDIA A100 GPU with 40GB of memory.

\section{Numerical Results}
\label{sec:numericals}
\subsection{Operator Learning for PDE Problems}
For operator learning tasks, we present detailed results on three different PDE problems. We select these problems based on their differing underlying physics to substantiate our findings. Additionally, these problems vary in dimension, domain, and operator formulation. For each problem, we modify the branch and trunk networks in accordance with our experimental setup. 
To verify the performance of the models, aside from the commonly used mean squared error (MSE) loss used in training, we also consider the popular $l_2$ relative error widely used for PDE problems, which reads
\[
l_2 \ \text{relative error} = \frac{\|\hat{u} - u\|_2}{\|u\|_2}.
\]
Here $u$ is the true solution and $\hat{u}$ is the prediction of the solution.

\subsubsection{Burgers' Equation}

\begin{figure}[ht!]
    \centering
    \includegraphics[width=\textwidth]{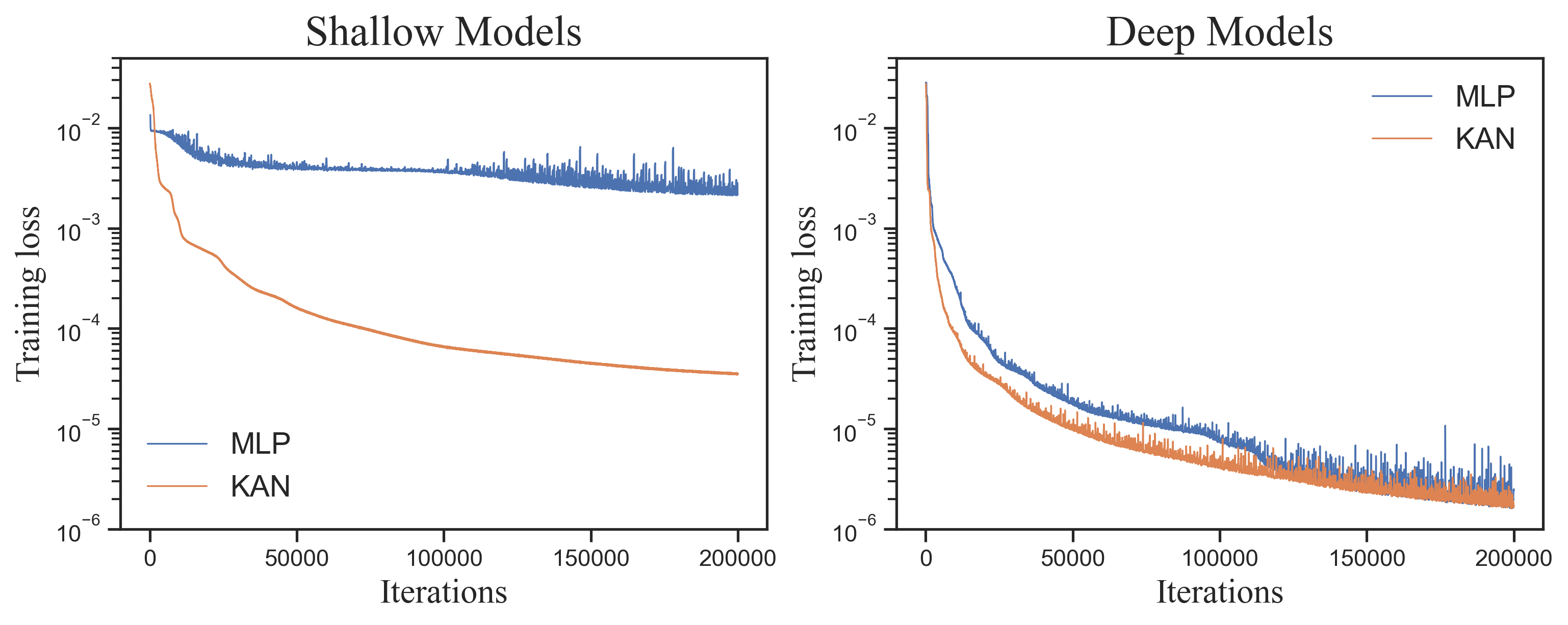}
    \caption{Training loss curves for Burger's equation problem. Left: results for the shallow network models. Right: results for the deep network models. The loss curves are smoothed using a moving average method for better representation. }
    \label{fig:burgers_training_loss}
\end{figure}

The first problem we consider is the 1-dimensional wave propagation described by Burger's equation, which serves as a common benchmarking problem~\citep{li2021fourierneuraloperatorparametric, ingebrand2024basistobasisoperatorlearningusing, kag2024learninghiddenphysicsparameters} for different operator learning methods. The PDE is described as follows: 
\begin{equation}
    \frac{\partial u}{\partial t}(x,t) + u\frac{\partial u}{\partial x}(x,t)= \nu \frac{\partial^2 u}{\partial^2 x}(x,t) 
\end{equation}
over the domain $\Omega$: $(x,t) \in [0,1]^2$,
and subject to periodic boundary conditions and some initial conditions. To set up the operator learning problem, we consider fixed viscosity $\nu=0.01$; the initial conditions $u(x,0)$ are randomly sampled from a Gaussian random field, and the corresponding solutions are solved over the domain using a finite difference method. The resulting data set contains a total of $2500$ different examples. 
We aim to recover the operator $\mathcal{G}: u(x,0) \rightarrow u(x,t)$ for $(x,t) \in \Omega$. 

We split our experiment into two parts: in the case of the shallow network with only one hidden layer, for the branch network, we select a hidden size of $1000$ for the MLP and $203$ for the KAN model, given the initial condition's discretization. Similarly, for the trunk network, we use a hidden size of $1000$ for the MLP and $5$ for the KAN since the input dimension is $2$. We choose a total of $100$ basis functions according to the results from \citep{ingebrand2024basistobasisoperatorlearningusing}. Both the shallow KAN network and shallow MLP network are trained to 25,000 epochs using an Adam optimizer with a learning rate of 1e-4.
For experiments on deep networks, we set up our architecture for the MLP identical to the ones used in \citep{kag2024learninghiddenphysicsparameters}, which use $7$ layers for their model with a hidden size of $128$. For the KAN model we implement our model following the experimental setup provided in \citep{shukla2024comprehensivefaircomparisonmlp}, where a model with $4$ layers and hidden size $100$ is used. Note here that the mentioned architectures are applied to both the branch network and the trunk network. 
We train the models using the Adam optimizer with a starting learning rate of $1e-4$, and we gradually decrease the learning rate for the MLP case to ensure optimal results.

We observe significantly better accuracy using the KAN architecture for shallow networks. The KAN model achieves over a magnitude lower loss and similar improvement in terms of $l_2$ relative error, suggesting the representation advantage of the KAN architecture in shallow networks.

\begin{table}[ht]
    \centering
    \caption{Test error of shallow and deep network models for the Burgers example.}
    \begin{tabularx}{\textwidth}{l X X }
    \toprule
    \textbf{Model} & \textbf{MSE Error} & \textbf{Rel. \( l_2 \) Error} \\ \midrule
    MLP (shallow) & 1.88e-3 & 23.1\%\\ 
    KAN (shallow)  & 3.79e-5 & 2.45\%\\  \midrule
    MLP (deep) & 2.82 e-6 & 0.89\% \\ 
    KAN (deep) & 5.95 e-6 & 0.82\% \\ 
    \bottomrule
    \end{tabularx}
    \label{tab:burgers_error}
\end{table}

For the comparison of deep networks, both the KAN model and the MLP model show similar performance, with only $0.07\%$ difference in relative $l_2$ error. While the KAN model converges at a faster rate than the MLP counterpart, the training loss after $200,000$ iterations are similar.  Detailed results are presented in \cref{tab:burgers_error} and  \cref{fig:burgers_training_loss}. 

\begin{figure}[ht!]
    \centering
    \includegraphics[width=\textwidth]{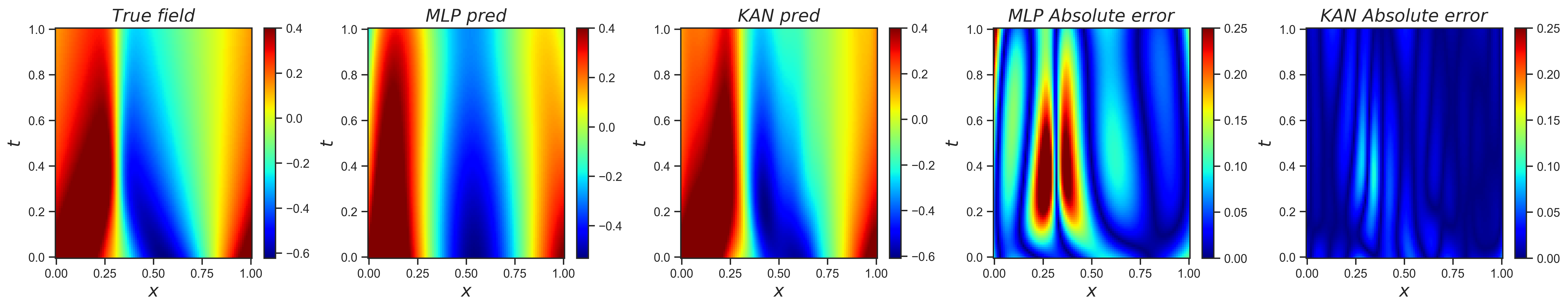}
    \caption{prediction of shallow network models for the Burgers example. From left to right: True solution, MLP prediction, KAN prediction, absolute error for the MLP prediction, absolute error for the KAN prediction.}
    \label{fig:burgers_shallow_pred}
\end{figure}

\begin{figure}[ht!]
    \centering
    \includegraphics[width=\textwidth]{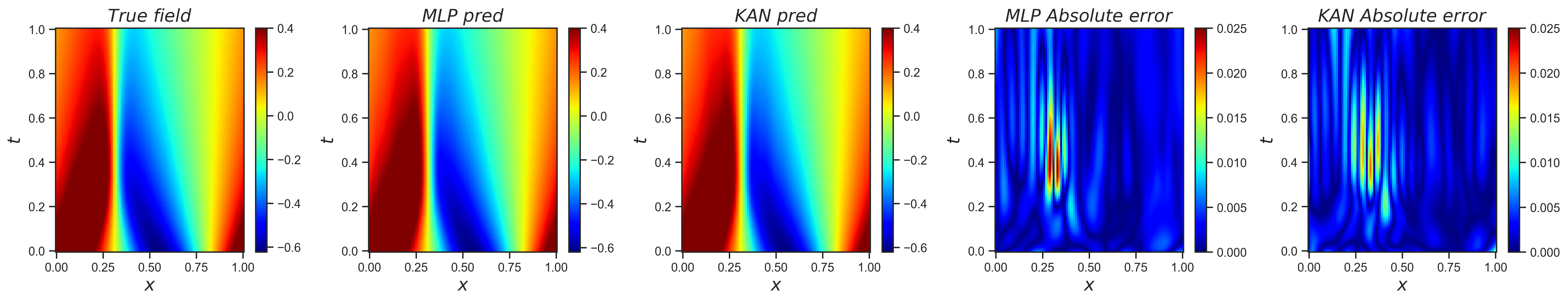}
    \caption{prediction of deep network models for the Burgers example. From left to right: True solution, MLP prediction, KAN prediction, absolute error for the MLP prediction, absolute error for the KAN prediction.}
    \label{fig:burgers_deep_pred}
\end{figure}

Finally, we present a sample of predictions and ground truth solutions from the test set, showing both the MLP and KAN predictions as well as errors versus the ground truth. We choose the presented sample as having close to the mean test-error in an $l_2$ sense. The shallow network case is presented in \cref{fig:burgers_shallow_pred} and the deep network case in \cref{fig:burgers_deep_pred}.

\subsubsection{1D Darcy Flow}

\begin{figure}[h!]
    \centering
    \includegraphics[width=\textwidth]{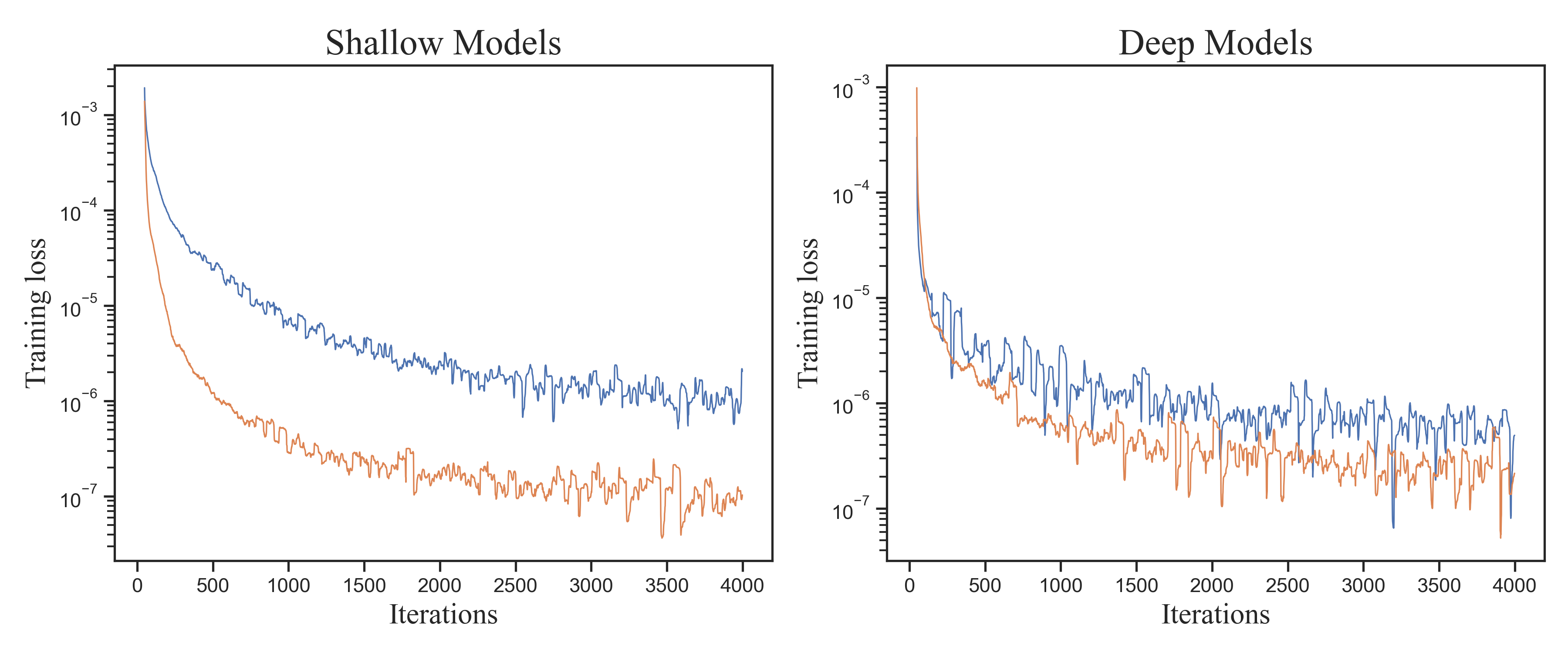}
    \caption{Training loss curves for 1D Darcy problem. The loss curves are smoothed using a moving average method for better representation.}
    \label{fig:1d_darcy_training_loss}
\end{figure}

We next consider the nonlinear 1D Darcy equations described by 
\begin{equation}
    \frac{du}{dx}(-\kappa(u(x))\frac{du}{dx}) = s(x), \quad x \in [0,1],
\end{equation}
with solution-dependent permeability given by $$\kappa( u(x) ) = 0.2 + u^2(x)$$ and the source function $s(x)$ generated from a Gaussian random field. Homogeneous Dirichlet boundary conditions $u = 0$ are considered at the domain boundaries. The problem definition as well as the data set comes from \citep{ingebrand2024basistobasisoperatorlearningusing}. For each sampled source term the solution is then obtained using a finite difference method, resulting in a dataset split into 800 training and 150 test examples.
The goal is to learn the operator $\mathcal{G}: s(x) \rightarrow u(x), x \in [0,1]$ for the 1D system.

For this problem, we select $p=100$ to be the latent dimension of our DeepONet models, consistent with other experiments. Given that the source function $s(x)$ is discretized over a uniform grid of size 50, for the shallow network setting, we set the hidden size of the branch net to $1000$ for the MLP and $101$ for the KAN model to process the source function; for predicting the solution, the trunk networks have hidden sizes $1000$ for the MLP and $3$ for the KAN model. We train with a batch size of $32$ for $4000$ epochs. The shallow MLP requires a learning rate of $1e-5$ for best results,  whereas the KAN model achieves optimal results with a learning rate of $1e-4$. Recall that the shallow KAN network has a grid size of 20 in the trunk net, as we achieve optimal results under such setting. 

\begin{figure}[ht]
    \centering
    \includegraphics[width=1.0\textwidth]{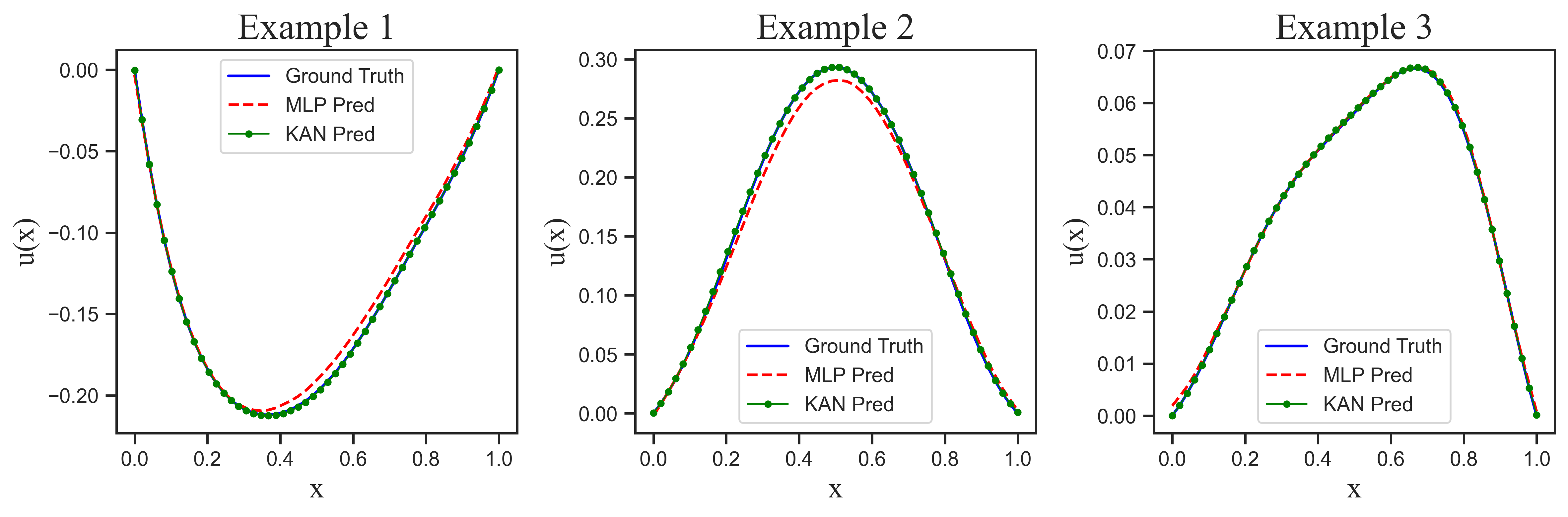}
    \caption{Prediction results of the shallow network models for the 1D Darcy problem.}
    \label{fig:1d_darcy_shallow_test_results}
\end{figure}

We first present results for the shallow experiments, as shown in \cref{fig:1d_darcy_training_loss} and \cref{tab:1d_darcy_error_comparison}. In general, we observe lower training loss and less approximation error in shallow KANs compared to the shallow MLP networks in the branch and trunk nets. Examples of approximation results are presented in \cref{fig:1d_darcy_shallow_test_results}, where it is clear that the KAN model is more accurate in prediction.

Our experiments comparing deep neural networks follow the setup provided in \citep{ingebrand2024basistobasisoperatorlearningusing} closely: our deep MLP networks use 4 layers with a hidden size of $256$ for both branch and trunk networks. The KAN models use 4 layers with hidden size $100$ for both networks. Both networks are trained using an Adam optimizer with $1e-4$ learning rate, with the batch size and number of epochs consistent with the shallow training case. 

\begin{figure}[ht]
    \centering
    \includegraphics[width=1.0\textwidth]{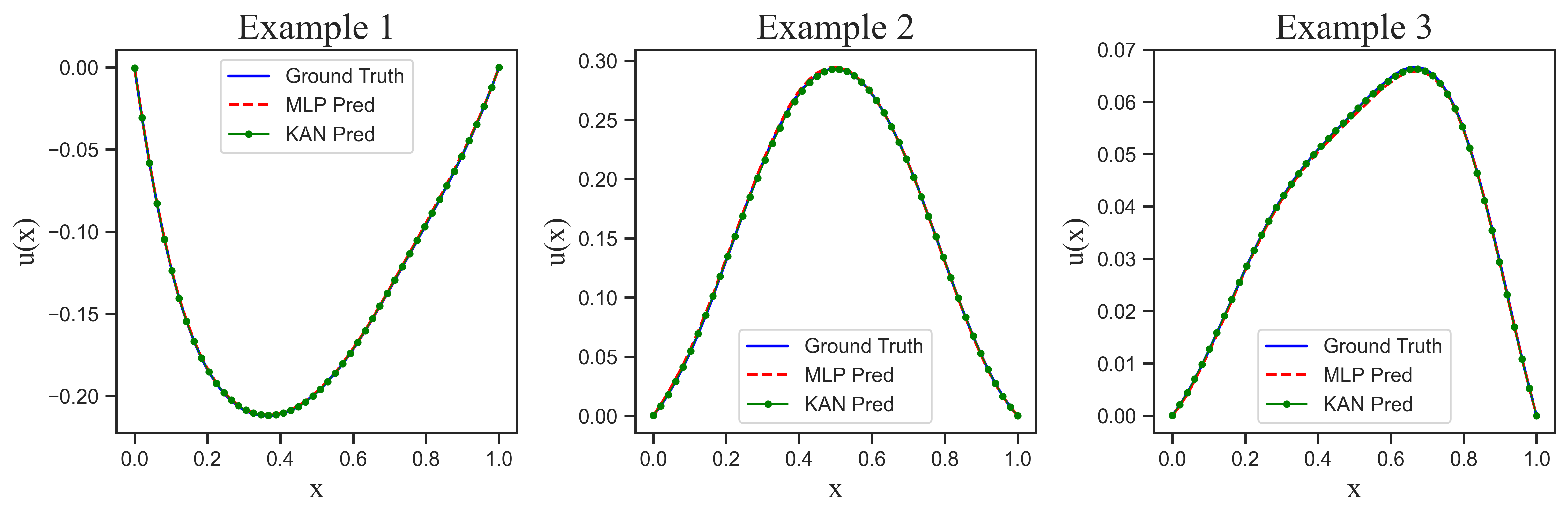}
    \caption{Prediction results of the deep network models for the 1D Darcy problem.}
    \label{fig:1d_darcy_deep_test_results}
\end{figure}

In the deep case, as presented in \cref{fig:1d_darcy_training_loss} and \cref{tab:1d_darcy_error_comparison}, both networks are able to achieve small training loss magnitudes in a stable fashion. In terms of relative $l_2$ error, the KAN models show an advantage over the MLP model, though the difference is much smaller than that of the shallow case. 
Given the smaller size of the problem, we train the deep KAN network with a trunk net grid size of 20 as well, in order to ensure best performance of the network. We observe no meaningful improvement over the baseline setting; nevertheless, we present the result in \cref{tab:1d_darcy_error_comparison}. This suggests that hyperparameter tuning for deep KAN models remains a challenge in many cases. 
Examples for model predictions are shown in \cref{fig:1d_darcy_deep_test_results} where both models can achieve high accuracy.

Finally, we include a summary of $l_2$ relative errors for each of the networks in~\cref{tab:1d_darcy_error_comparison}, to summarize the differences in performance. We also use the example to test the trained models' robustness to noise: 
models trained on the original data are tested on new datasets of varying noise levels of $1\%$, $5\%$, and $10\%$. To create these datasets, Gaussian noise of different magnitude is added to the input functions and corresponding outputs are then generated using a finite difference solver to construct a ``noisy dataset''; the models trained on the original noiseless data are then evaluated on these noisy outputs, as has been done similarly in \citep{shukla2024comprehensivefaircomparisonmlp}. 
From \cref{tab:1d_darcy_error_comparison} both MLP and KAN show similar performances under different magnitudes of noise, indicating that the KAN models are robust to noisy data, which can be common in physics-based applications.

\begin{table}[h!]
    \centering
    \small

    \begin{tabularx}{\textwidth}{l X X X X } 
        \toprule
        \textbf{Model} & \textbf{Rel. $l_2$ Error} & \textbf{1\% Noise} & \textbf{5\% Noise} & \textbf{10\% Noise} \\ \midrule
        MLP (shallow) & $2.89\%$ & $3.13\%$ & $5.58\%$ & $9.42\%$ \\ 
        KAN (shallow) & $0.39\%$ & $0.99\%$ & $4.34\%$ & $8.61\%$ \\ \midrule
        MLP (deep) & $1.26\%$ & $1.66\%$ & $4.67\%$ & $8.80\%$ \\
        KAN (deep) & $0.42\%$ & $1.00\%$ & $4.35\%$ & $8.61\%$ \\ 
        \bottomrule
    \end{tabularx}
    \caption{Error statistics for 1D Darcy models.}
    \label{tab:1d_darcy_error_comparison}
\end{table}

\subsubsection{2D Darcy Flow}

\begin{figure}[ht]
    \centering
    \includegraphics[width=\textwidth]{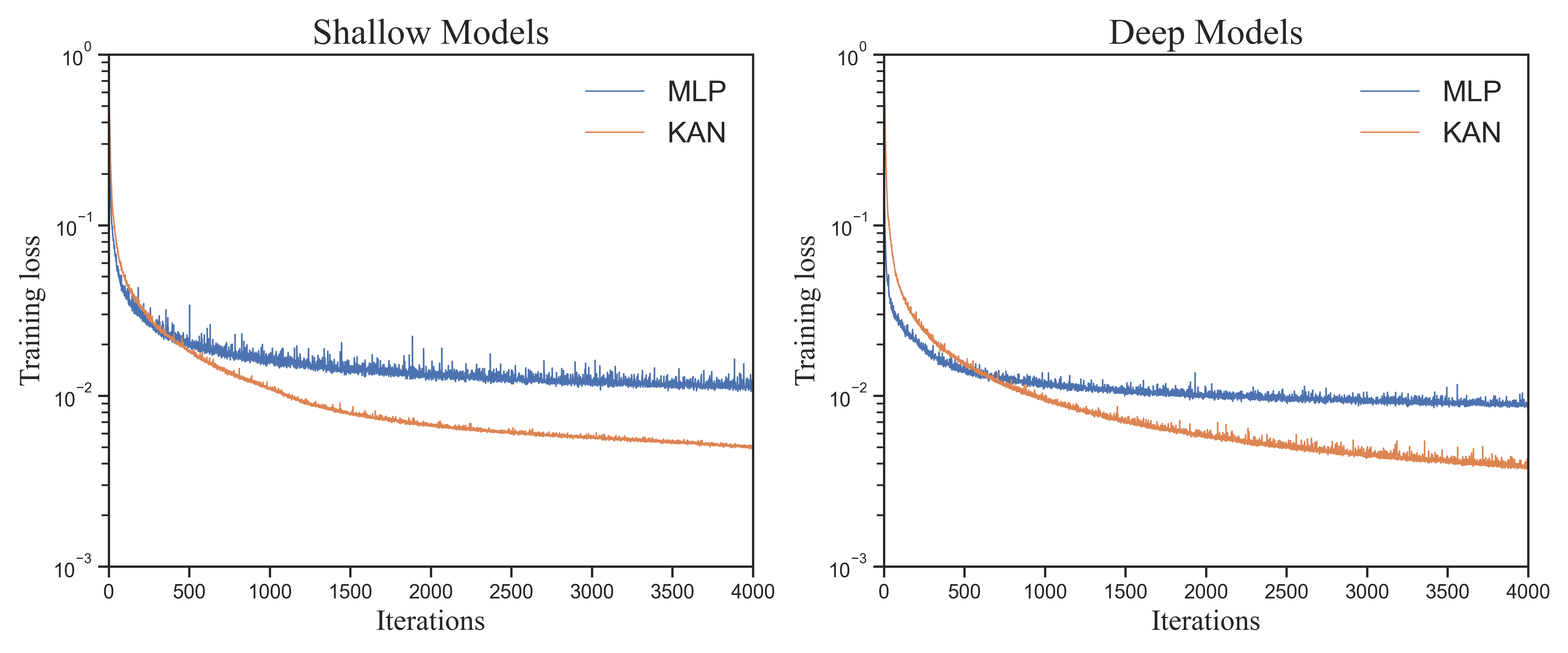}
    \caption{Training and test loss curves for the 2D Darcy problem. The loss curves are smoothed using a moving average method for better representation.}
    \label{fig:2d_darcy_training_loss}
\end{figure}

We consider a 2D Darcy flow problem, a relatively complex benchmark problem also used in \citep{ingebrand2024basistobasisoperatorlearningusing, kahana2023geometry}. The problem is defined as
\begin{equation}
    \begin{aligned}
        \nabla \cdot (k(\boldsymbol{x}) \nabla u(\boldsymbol{x})) + f(\boldsymbol{x}) &= 0, \quad \boldsymbol{x} = (x,y) \in \Omega := (0,1)^2 \setminus [0.5,1)^2, \\
        u(\boldsymbol{x}) &= 0, \quad \boldsymbol{x} \in \partial \Omega.
    \end{aligned}
\end{equation}
Here, $k(\boldsymbol{x})$ is the permeability field, $u(\boldsymbol{x})$ the solution hydraulic head, and $f(\boldsymbol{x})$ the spatially varying force vector. Note here that instead of a regular domain we have an L-shaped domain $\Omega$, as well as two different input functions for the system, both of which add to the difficulty of the problem.
The goal is to learn the nonlinear operator mapping the permeability field and source vector to the hydraulic flow pressure i.e. to learn $\mathcal{G}: [k(\boldsymbol{x}), f(\boldsymbol{x})] \rightarrow u(\boldsymbol{x}) \quad \forall \boldsymbol{x} \in \Omega$.

We use the data provided in \citep{ingebrand2024basistobasisoperatorlearningusing} which discretizes both input functions over a uniform $31\times 31$ grid over the square domain with  the top right portion padded with zeros. This results in an input dimension of $961$ to the branch network, a significantly larger input size than earlier problems. However, the solution space is defined on an irregular grid which is discretized with $450$ points.
The latent dimension is chosen to be $p=100$ as with other problems, based on \citep{ingebrand2024basistobasisoperatorlearningusing}. Notably, because our problem here processes two input source fields, our architecture is an augmented DeepONet with two branch networks, in which the outputs of each branch net are added together before combining with the trunk net output.
For all tests,  we normalize the dataset before splitting into training and test sets for optimal results and to avoid vanishing gradients. We use an Adam optimizer with learning rate of $1e-4$ for all tests. Our shallow MLP-based network uses hidden size $1000$. The shallow KAN network has two branch nets with hidden size of $1923$ and trunk net with hidden size $5$. 
The deep networks are configured in the same ways as in previous examples and \citep{ingebrand2024basistobasisoperatorlearningusing, shukla2024comprehensivefaircomparisonmlp} where we use a $4$ layer network for the MLP model with hidden size $256$ and a $4$ layer network for the KAN model with hidden size $100$. 

\begin{figure}[H]
    \centering
    \begin{minipage}{\textwidth}
        \centering
        \includegraphics[width=\textwidth]{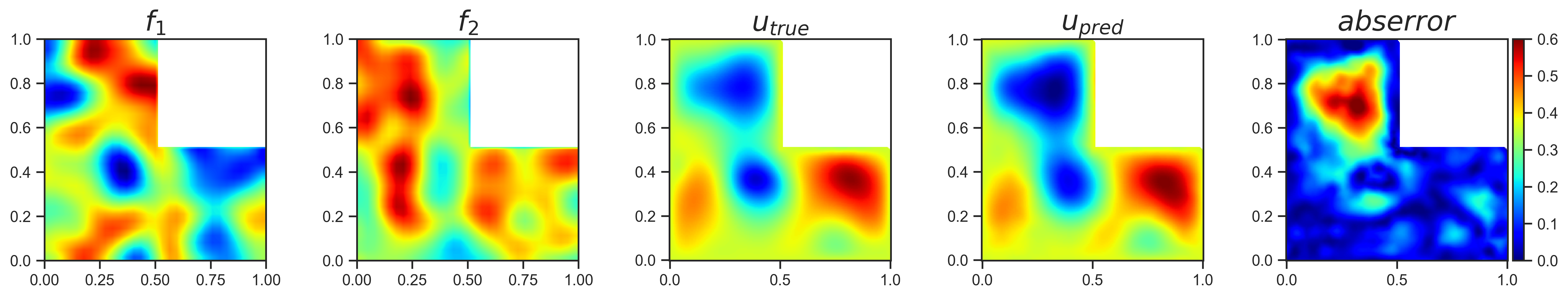}
        \caption*{(a) MLP}
        \label{fig:2d_darcy_shallow_MLP_result}
    \end{minipage}
    \hfill
    \begin{minipage}{\textwidth}
        \centering
        \includegraphics[width=\textwidth]{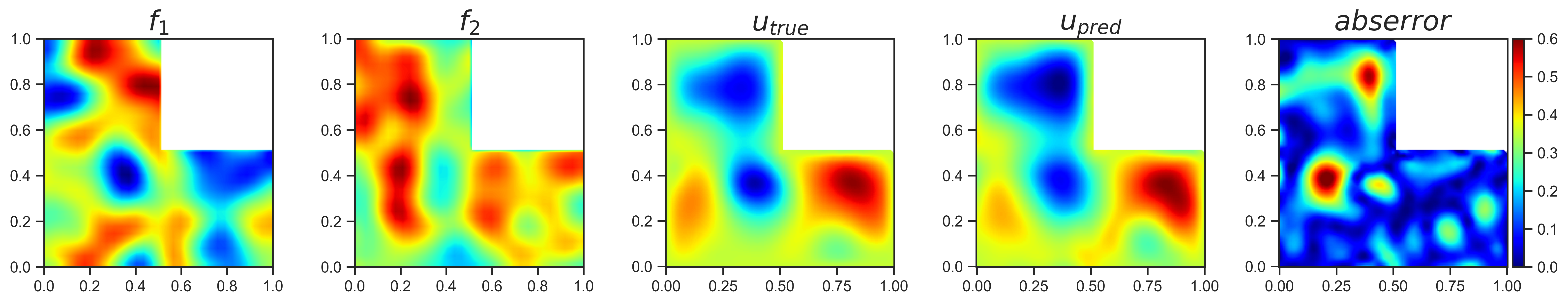}
        \caption*{(b) KAN}
        \label{fig:2d_darcy_shallow_kan_result}
    \end{minipage}
    \caption{Test results of shallow network models for the 2D Darcy problem. From left to right: input functions, true solution, model prediction and absolute error of the predicted solution.}
    \label{fig:2d_darcy_shallow_test_results}
\end{figure}

\begin{figure}[ht]
    \centering
    \subfloat[MLP]{
        \includegraphics[width=\textwidth]{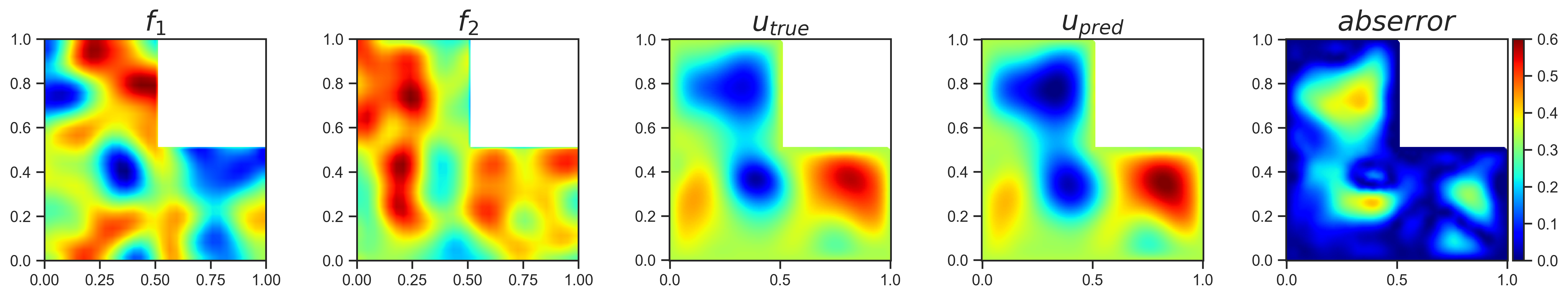}
        \label{fig:2d_darcy_deep_MLP_result}
    }
    
    \subfloat[KAN]{
        \includegraphics[width=\textwidth]{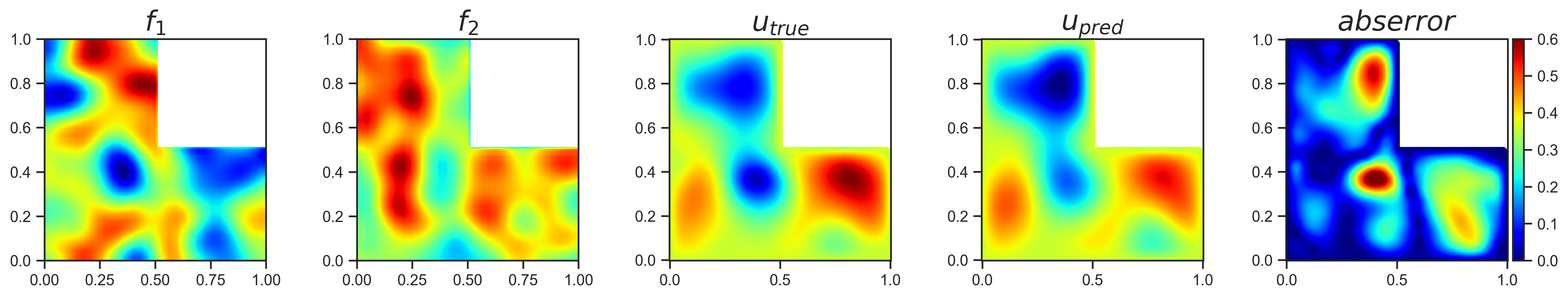}
        \label{fig:2d_darcy_deep_kan_result}
    }
    
    \caption{Test results of deep network models for the 2D Darcy problem. From left to right: input functions, true solution, model prediction and absolute error of the predicted solution.}
    \label{fig:2d_darcy_deep_test_results}
\end{figure}

We first present smoothed training loss curves below for both experiments in \cref{fig:2d_darcy_training_loss}, and detailed error results are listed in \cref{tab:2d_darcy_error_comparison}. 
We note that for this example in both cases the MLP model outperforms the KAN model, indicated by lower test loss and smaller relative $l_2$ error when evaluating the models. 
It is also important to point out that in training, the KAN models are able to achieve much lower training losses than the MLP-based models, despite worse performance on the test set. This suggests that the KAN models suffer much more from over-fitting than the MLP models. 
We attribute this to the added complexity of the problem with two different input functions and the relative sparsity of the available data to solve the problem. 
Furthermore, the high dimensionality of the input data, its irregular domain, and the imbalance between the branch and trunk net dimensions may contribute to the difficulty in learning the true operator. That said, we note that it is still evident that KAN architecture has a distinct advantage in representation capability, which coincides with earlier examples. 
We display examples and predictions from different models in \cref{fig:2d_darcy_shallow_test_results} and \cref{fig:2d_darcy_deep_test_results}.

\begin{table}
    \centering
    \small
    \begin{tabularx}{\textwidth}{l X X } 
        \toprule
        \textbf{Model} & \textbf{Rel. $l_2$ Error} & \textbf{MSE Error} \\ \midrule
        MLP (shallow) & $20.7\%$ & 3.59e-2 \\ 
        KAN (shallow) & $23.2\%$ & 4.18e-2 \\ \midrule
        MLP (deep) & $19.7\%$ & 3.62e-2 \\ 
        KAN (deep) & $23.3\%$ & 5.45e-2 \\ \bottomrule
    \end{tabularx}
    \caption{Error statistics for 2D Darcy models.}
    \label{tab:2d_darcy_error_comparison}
\end{table}

\subsubsection{Elastic Plate}

Finally, we consider a two-dimensional problem which we refer to as the elastic plate example. We consider a rectangular plate subjected to in-plane loading that is modeled as a problem of plane stress elasticity. The governing equation is given by
\[
\nabla \cdot \sigma + f(\boldsymbol{x}) = 0, \quad \boldsymbol{x} = (x, y) \in \Omega,
\]
\[
(u, v) = 0, \quad \forall \; x = 0,
\]
where \(\sigma\) is the Cauchy stress tensor, \(f\) is the body force, \(u\) and \(v\) represent the \(x\)- and \(y\)-displacement, respectively. In addition to that, \(E\) and \(\nu\) represent the Young modulus and the Poisson ratio of the material, respectively. The relation between stress and displacement can be described as
\[
\begin{bmatrix}
\sigma_{xx} \\
\sigma_{yy} \\
\tau_{xy}
\end{bmatrix}
=
\frac{E}{1 - \nu^2}
\begin{bmatrix}
1 & \nu & 0 \\
\nu & 1 & 0 \\
0 & 0 & \frac{1 - \nu}{2}
\end{bmatrix}
\begin{bmatrix}
\frac{\partial u}{\partial x} \\
\frac{\partial v}{\partial y} \\
\frac{\partial u}{\partial y} + \frac{\partial v}{\partial x}
\end{bmatrix}.
\]
We generate the loading conditions \(f(\boldsymbol{x})\) applied to the right edge of the plate from a Gaussian random field. We aim to recover the mapping \(\mathcal{G}: f(\boldsymbol{x}) \rightarrow [u(\boldsymbol{x}), v(\boldsymbol{x})]\) given the different random boundary loads.
We adapt the dataset for the problem from \citep{ingebrand2024basistobasisoperatorlearningusing, goswami2022deep}, which has 1850 training and 100 test examples for this problem.
As in previous examples, we use a DeepONet architecture with the latent dimension chosen as $p=100$. The input dimension to the branch net is 101, based on the sampling of the dataset. For all experiments, we use an Adam optimizer with learning rate $1e-4$ and normalize the data before passing them through the networks. Our shallow MLP-based DeepONet uses a hidden size of $1000$, and the shallow KAN model has a branch net of hidden size $203$ and trunk net of hidden size $5$. Also similar to earlier examples, we increase the spline grid size to $20$ in the trunk net of the shallow KAN network to improve the representational ability of the model. The deep MLP-based networks use $4$ layers of size $256$, and the deep KAN-based networks use $4$ layers of size $100$, with a smaller default grid size of $5$. We obtain optimal results under the aforementioned setup. 

\begin{figure}
    \centering
    \includegraphics[width=1.0\linewidth]{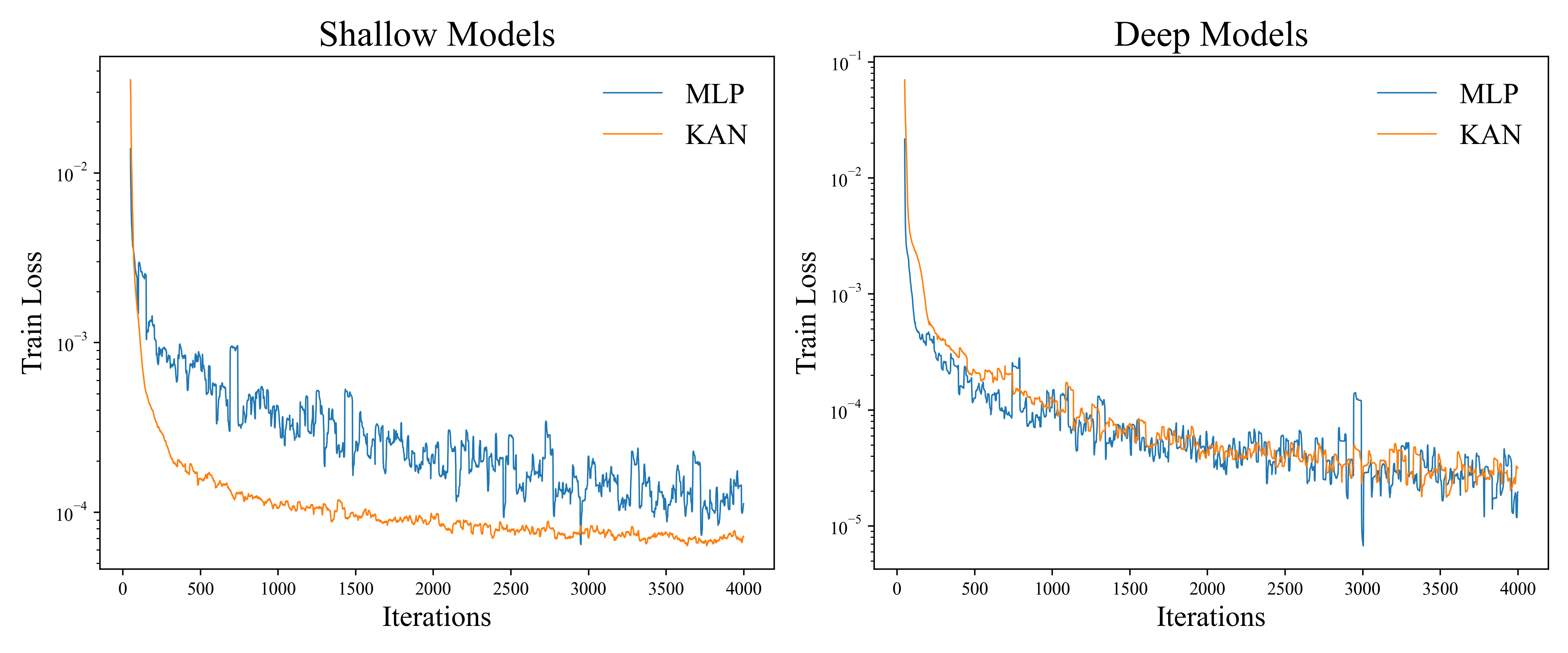}
    \caption{Loss curves for the elastic plate problem. Curves are smoothed using a moving average method for better presentation.}
    \label{fig:elastic_plate_loss}
\end{figure}

\begin{figure}[h]
    \centering
    \subfloat[Shallow network models.]{%
        \includegraphics[width=0.48\linewidth]{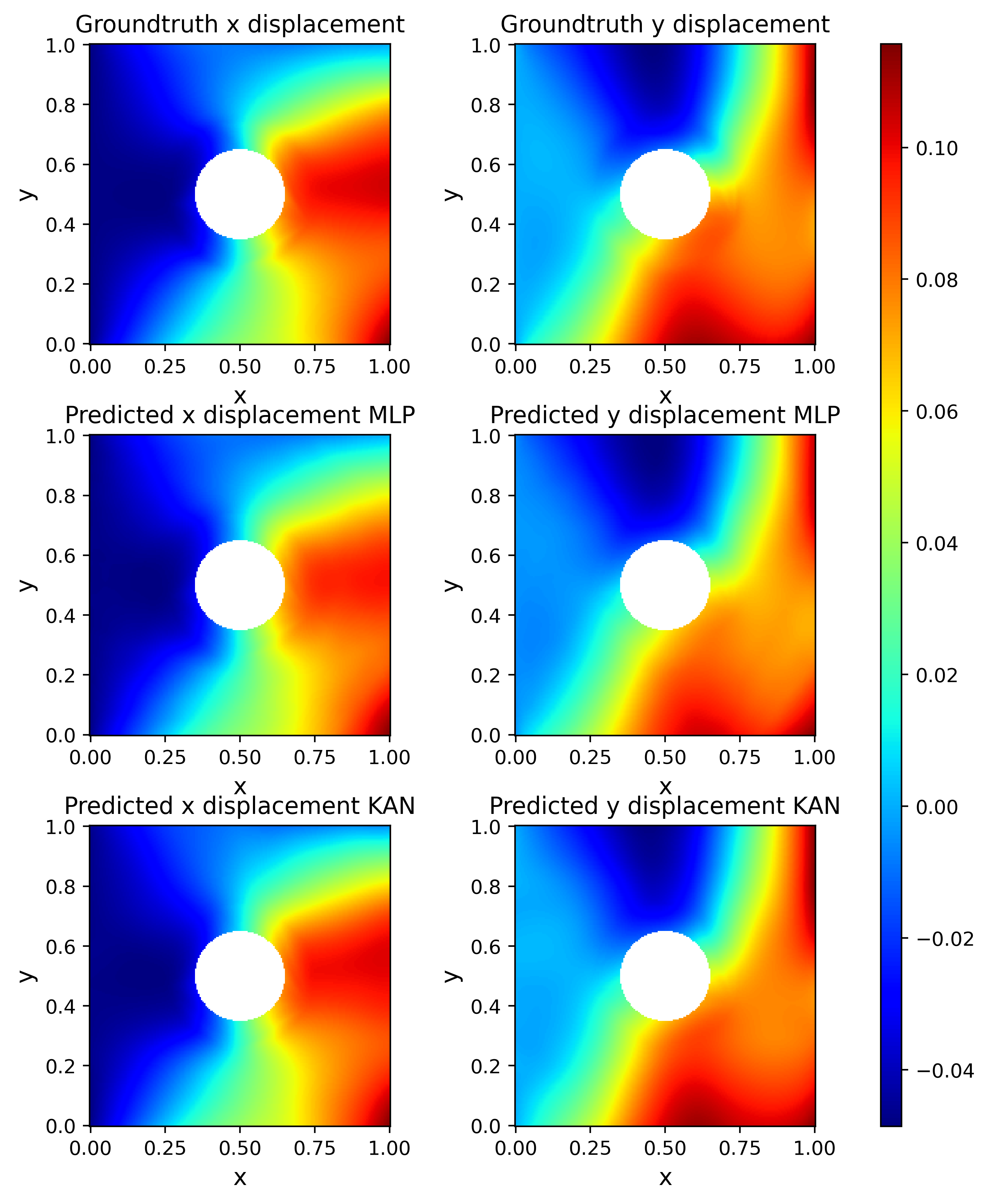}
    }
    \hfill
    \subfloat[Deep network models.]{%
        \includegraphics[width=0.48\linewidth]{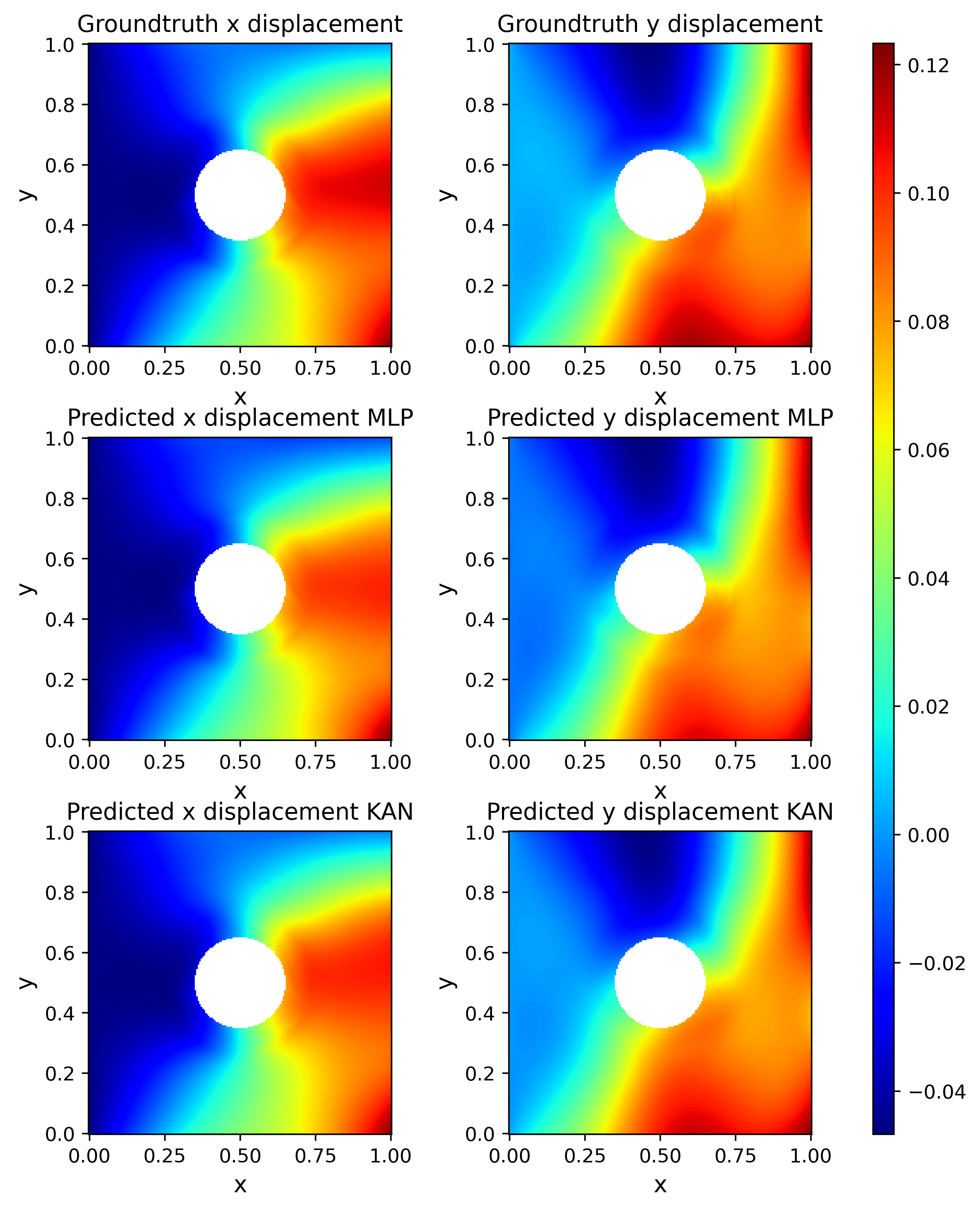}
    }
    \caption{Test set results for the elastic plate problem. Left: results for shallow neural networks. Right: results for deep neural networks.}
    \label{fig:elastic_plate_test_results}
\end{figure}

We display the training and final test results in \cref{fig:elastic_plate_loss} and \cref{tab:elastic_plate_error_comparison}. Similarly to other examples,  we observe that a shallow KAN model can outperform its MLP counterpart in both convergence speed and overall test accuracy. On the other hand, as we move to deep networks, the differences between models are not as significant; while for the example we still see the KAN model achieving a lower test error, the improvement from shallow networks is minimal.  Qualitative results are shown in \cref{fig:elastic_plate_test_results}.

\begin{table}
    \centering
    \small
    \begin{tabularx}{\textwidth}{l X X } 
        \toprule
        \textbf{Model} & \textbf{MSE Error} & \textbf{Rel. $l_2$ Error}\\ \midrule
        MLP (shallow) & 3.21e-4 & $2.93\%$ \\ 
        KAN (shallow) & 9.70e-5 & $1.35\%$ \\ \midrule
        MLP (deep) & 3.94e-5 & $1.1\%$ \\ 
        KAN (deep) & 5.71e-5 & $0.84\%$ \\ \bottomrule
    \end{tabularx}
    \caption{Error performance for elastic plate models.}
    \label{tab:elastic_plate_error_comparison}
\end{table}

\paragraph{Lipschitz Continuity Test}
To ensure our testing is extensive, we also conduct additional experiments using the elastic plate example. First, we test to verify the Lipschitz continuity of the learned operator model. Recall that for normed vector spaces $\mathcal{U}$ and $\mathcal{V}$, we define the Lipschitz continuity of the operator $\mathcal{G}$ if there exists a constant $L \geq 0$ such that
\[
\| \mathcal{G}(u_1) - \mathcal{G}(u_2) \|_V \leq L \| u_1 - u_2 \|_U \quad \text{for all } u_1, u_2 \in \mathcal{U}.
\]
We refer to the smallest such constant \( L \) as the Lipschitz constant. The Lipschitz constant serves as a useful tool in determining the robustness of a learned model: as such, we evaluate this using the learned models over the test data set dataset consisting of inputs $\{ u_i \}_{i=1}^{N}$ under the formula
\[
L_{\mathcal{G}} = \max_{\substack{i, j \in \{1, \dots, N\} \\ i \neq j}} \frac{\|\mathcal{G}(u_i) - \mathcal{G}(u_j)\|}{\|u_i - u_j\|},
\]
here we use the $2$-norm for both input and output space. 
We present the results in \cref{tab:lipschitz_elastic_plate} for all four models as well as the ground truth results from the data set. We notice that all the learned models show similar Lipschitz constant to the ground truth, indicating that the learned models are robust across different test functions.

\begin{table}[h]
    \centering
    \small
    \renewcommand{\arraystretch}{1.5} 
    \begin{tabularx}{\textwidth}{
        >{\centering\arraybackslash}m{2.5cm} 
        >{\centering\arraybackslash}X 
        >{\centering\arraybackslash}X 
        >{\centering\arraybackslash}X 
        >{\centering\arraybackslash}X 
        >{\centering\arraybackslash}X }
        \toprule
         & \textbf{Shallow MLP} & \textbf{Shallow KAN} & \textbf{Deep MLP} & \textbf{Deep KAN} & \textbf{Ground Truth} \\ \midrule
        \textbf{Lipschitz Constant} & $4.025$ & $4.033$  & $4.014$ & $4.013$ & $4.008$ \\ 
        \bottomrule
    \end{tabularx}
    \caption{Lipschitz constant of learned operators and ground truth.}
    \label{tab:lipschitz_elastic_plate}
\end{table}

\paragraph{Out-Of-Distribution Predictions}
To further assess the robustness of the already trained models, we evaluate their performance on unseen data that is significantly distinct from both the original training and original testing datasets. Our aim is to verify whether the KAN and MLP models could effectively handle these out-of-distribution (OOD) predictions without a substantial increase in errors. We test this using the elastic plate example. 

\begin{figure}
    \centering
    \includegraphics[width=\linewidth]{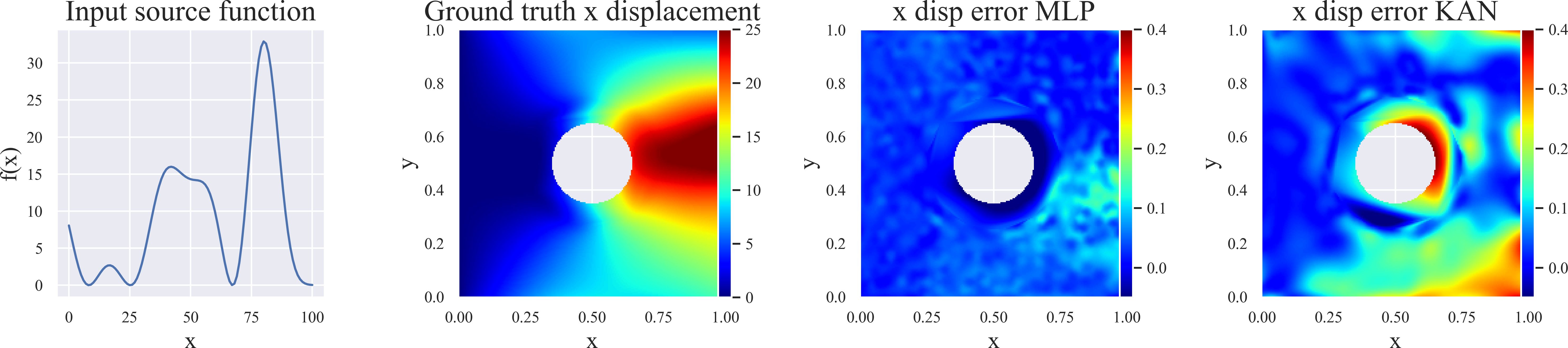}
    \caption{OOD example comparing deep MLP and KAN-based models.}
    \label{fig:ood-final}
\end{figure}

To test this, we take the input functions from the test dataset and increase their magnitude by a factor of 10. This results in input functions that are substantially different from the original training and test examples. The corresponding solutions to the inputs are then computed for reference. In total, 10 examples are generated for this test. We calculate the prediction results from all trained models and compare them against the ground truth solutions; numerical results are presented in \cref{tab:elastic_plate_ood}, and we include an example qualitative plot in \cref{fig:ood-final}.

We observe that while MLP-based models exhibit limited performance degradation on OOD data, KAN-based models show a significantly larger increase in error. We attribute this to the nonlinear parameterization heavily used in KAN architectures, suggesting that although KANs can outperform on in-distribution data, they face disadvantages when generalizing to out-of-distribution examples.

\begin{table}
    \centering
    \small
    \begin{tabularx}{\textwidth}{l *{4}{>{\centering\arraybackslash}X}} 
        \toprule
        \textbf{Metric} & \textbf{MLP\newline(shallow)} & \textbf{KAN\newline(shallow)} & \textbf{MLP\newline(deep)} & \textbf{KAN\newline(deep)} \\ \midrule
        Rel. $l_2$ Error & $2.17\%$ & $3.68\%$ & $0.88\%$ & $3.51\%$ \\
        \bottomrule
    \end{tabularx}
    \caption{Relative $l_2$ error on OOD test data for elastic plate models.}
    \label{tab:elastic_plate_ood}
\end{table}

To briefly summarize, our tests across multiple examples demonstrate that KAN-based operator networks significantly outperform MLP-based models when implemented as shallow neural networks, while the performance gap diminishes with deeper networks. 
These findings motivate further exploration of applications in scientific machine learning that benefit from the expressiveness and effectiveness of KAN in simple, shallow architectures, such as graph neural network-based physical simulators. 

\subsection{Particle Dynamics}
\label{sec:GNS_results}

Finally, we focus on the problem of particle dynamics that are to be learned with a graph network based simulator (GNS). Namely, the goal is to learn local particle interactions using their observed historical data, such that the learned simulator can predict future trajectories over long time horizons, potentially in new domains with obstacles, different initial particle positions or velocities than the ones previously seen during training.

\begin{figure}
    \centering
    \begin{subfigure}[b]{0.49\textwidth}
        \centering
        \includegraphics[width=\textwidth]{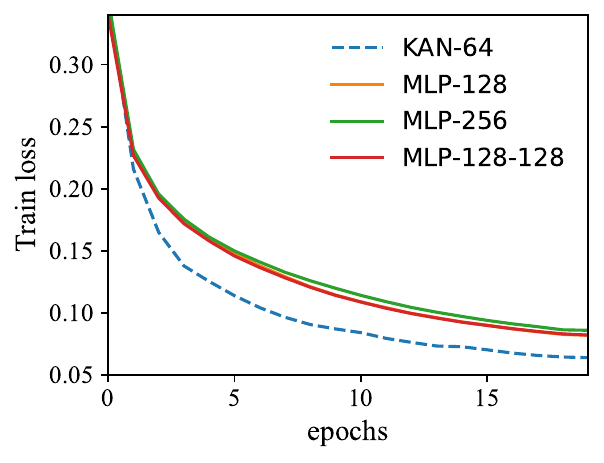}
        \caption{Sand}
    \end{subfigure}
    \begin{subfigure}[b]{0.49\textwidth}
        \centering
        \includegraphics[width=\linewidth]{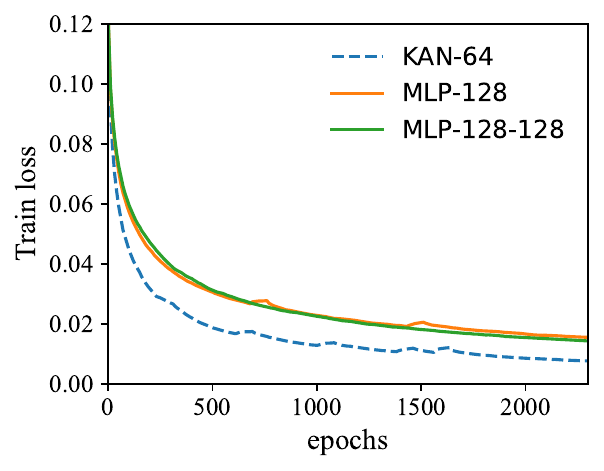}
        \caption{Water}
    \end{subfigure}
    \caption{Comparison of MLP and KAN based GNS models' training loss. For the \texttt{Sand} problem (left), models are trained for 19 epochs (2 million training steps), while for the \texttt{WaterDropSample} problem (right), models are trained for 3001 epochs (2 million training steps).}
    \label{fig:GNS_sand_losses}
\end{figure}

For a comprehensive comparative analysis, we construct and test four different GNS models by representing $\mathbf{f_\theta}$ and $\mathbf{g_\theta}$ in Eq. \eqref{eq:GNS_equations} with either shallow KAN or MLP of various sizes. We refer the reader to the discussion in \cref{sec:GNS_intro} on our motivation to test GNS represented with shallow KANs. Following the encoder, we remark that the information in GNS resides in a high-dimensional latent space as given by the node and edge embeddings dimensions. In this high-dimensional representation, shallow networks are sufficient to process the information effectively. Therefore, to benchmark performance of GNS represented with KAN, we construct GNS with three different yet shallow MLP configurations; a single hidden layer MLP of size $128$ (referred in \cref{fig:GNS_sand_losses} as \texttt{MLP-128}), single hidden layer MLP of size $256$ (\texttt{MLP-256}), and MLP with two hidden layers of sizes $128$ each (\texttt{MLP-128-128}). We compare the accuracy and stability of these models against a fourth GNS model represented with a single hidden layer KAN of size $64$ (\texttt{KAN-64}). We choose the hidden layer size for three reasons: it is less than
$2n+1$ where $n$ is the input dimension, it is sufficient to capture particle dynamics and meets the maximum memory limit of $40$ GB of our compute budget.

\begin{table}[h]
    \centering
    \begin{tabular}{lcccc}
        \toprule
        \cmidrule(lr){2-5}
        GNS model & \texttt{KAN-64} & \texttt{MLP-128} & \texttt{MLP-128-128} & \texttt{MLP-256} \\
        \midrule
        \texttt{Sand} & 2.36e-02 & 4.03e-02 & 5.05e-02 & 4.91e-02 \\
        \texttt{WaterDropSample} & 1.26e-01 & 9.16e-02 & 1.42e-01 & --- \\
        \bottomrule
    \end{tabular}
    \caption{MSE Test errors of GNS models on \texttt{Sand} and \texttt{WaterDropSample} dataset.}
    \label{tab:GNS_test_errors}
\end{table}

We illustrate the performance of our model across two physical simulations that exhibit different system dynamics over time. Namely, we use \texttt{WaterDropSample} and \texttt{Sand} datasets from \citep{sanchez2020learning,https://doi.org/10.17603/ds2-0phb-dg64} that describe the evolution of water droplets and granular soil particles, respectively, within a 2D container. Each dataset has up to $2000$ particles. The ground truth datasets are generated with a high-fidelity material point method (MPM), see \citep{kumar2019scalablemodularmaterialpoint}. We start GNS testing on \texttt{WaterDropSample} as it is smaller in size, and rely on the larger \texttt{Sand} dataset for a more extensive study. The training setup follows the standard GNS paradigm, where the model is trained to predict the next position of the particles given the previous five time steps. In all experiments, we use an Adam optimizer with initial learning rate of $0.0001$. We use a total of $10$ message passing layers. 

\begin{figure}[h]
    \centering
    \makebox[\textwidth]{\includegraphics[width=1.1\linewidth]{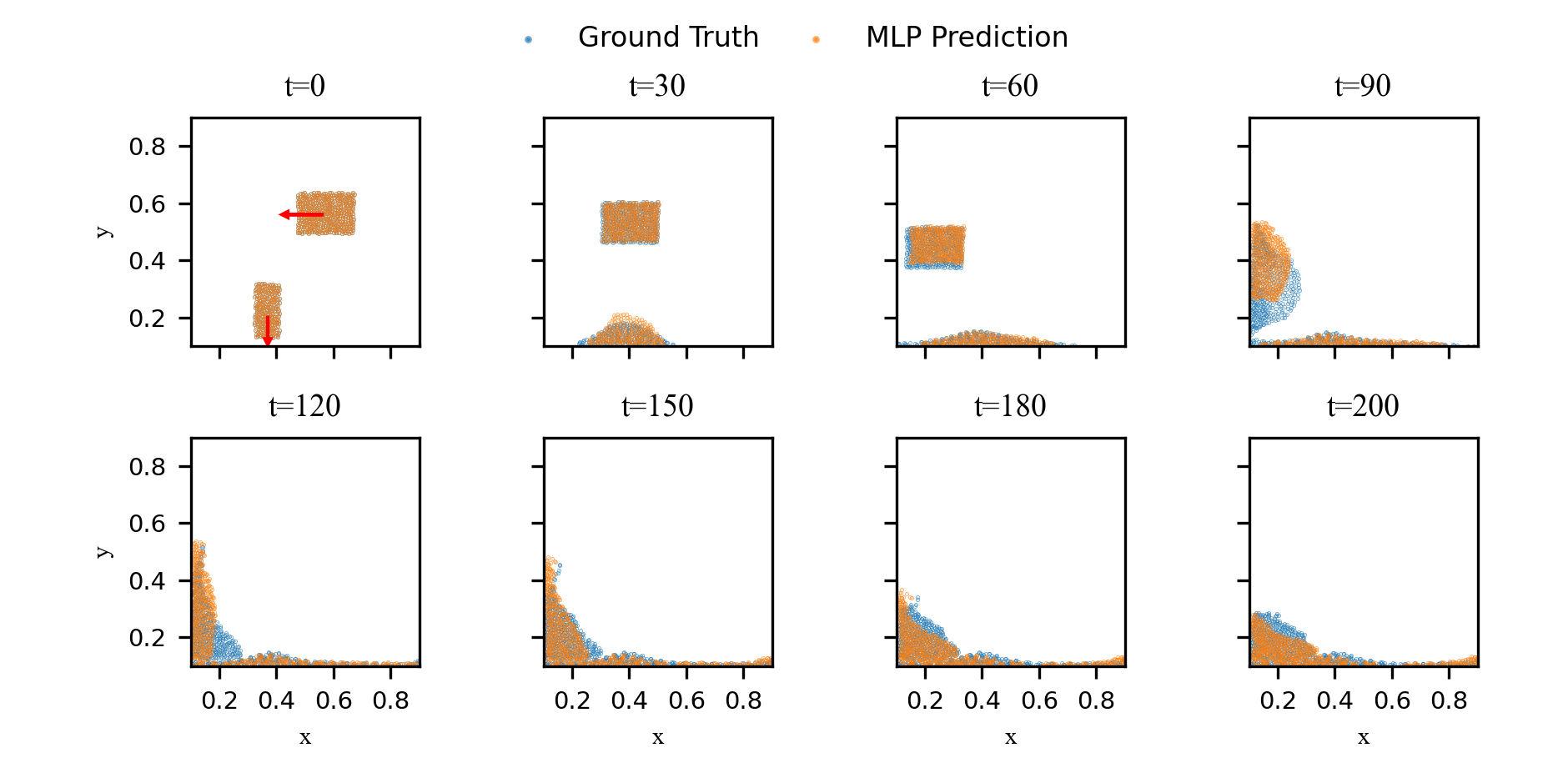}}
    \vspace{-1cm}
    \caption{Temporal snapshots comparing the qualitative performance of the \texttt{MLP-128-128} model's trajectory predictions with ground truth MPM data.}
    \label{fig:example_gns_mlp_rollout}
\end{figure}

\begin{figure}[h]
    \centering
    \makebox[\textwidth]{\includegraphics[width=1.1\linewidth]{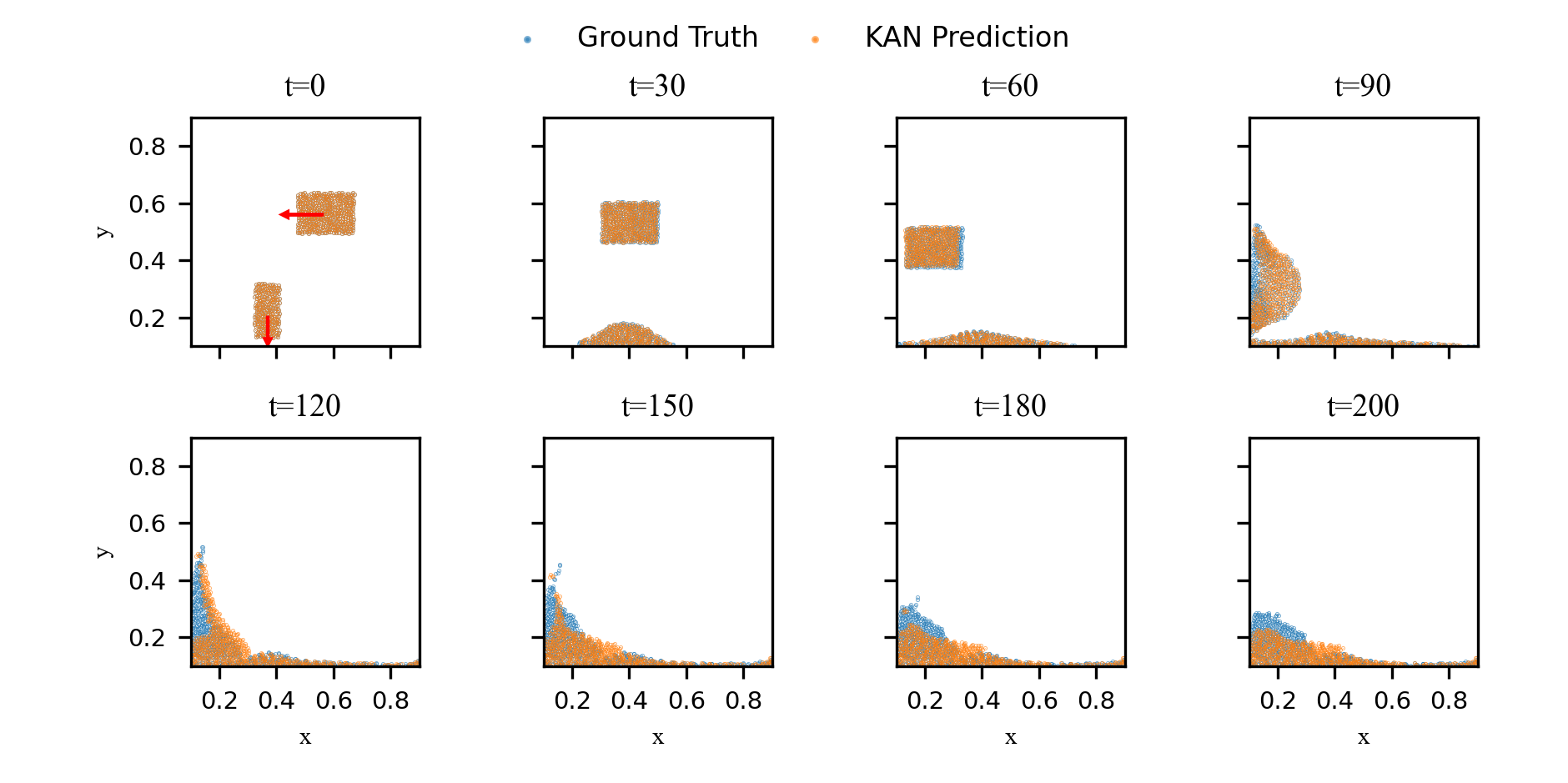}}
    \vspace{-1cm}
    \caption{Temporal snapshots comparing the qualitative performance of the \texttt{KAN-64} model's trajectory predictions with ground truth MPM data.}
    \label{fig:example_gns_kan_rollout}
\end{figure}

\begin{table}[h]
    \centering
    \begin{tabular}{lcccc}
        \toprule
        & \multicolumn{4}{c}{Time per 1000 Steps (H:MM:SS)} \\
        \cmidrule(lr){2-5}
        Experiment & \texttt{KAN-64} & \texttt{MLP-256} & \texttt{MLP-128-128} & \texttt{MLP-128} \\
        \midrule
        Time & 0:06:03 & 0:00:19 & 0:00:21 & 0:00:17 \\
        \bottomrule
    \end{tabular}
    \caption{Training time per 1000 steps for each implementation of the GNS models.}
    \label{tab:GNS_training_times}
\end{table}

We train all models for $2$ million training steps, corresponding to $19$ and $3001$ epochs for \texttt{Sand} and \texttt{WaterDropSample} problems respectively. The training loss curves, shown in \cref{fig:GNS_sand_losses}, indicate that \texttt{KAN-64} model achieves better accuracy compared to the MLP counterparts for both the \texttt{Sand} and the \texttt{WaterDropSample} problems. Test mean squared errors are presented in \cref{tab:GNS_test_errors}. We note that in terms of test errors, a strong case of overfitting is observed across all models and \texttt{KAN-64} do not outperform other models in the \texttt{WaterDropSample} problem. We attribute this overfitting to the much smaller number of training examples in the dataset, i.e. a larger number of samples are required to sufficiently capture complex particle dynamics and accurately predict in unseen test settings for the \texttt{WaterDropSample} problem. As the smaller \texttt{MLP-128} and \texttt{MLP-128-128} models show overfitting, we do not further test the bigger \texttt{MLP-256} model on the \texttt{WaterDropSample} problem. Testing a model with higher capacity would likely exacerbate the overfitting issue, as increased parameters would enable the network to memorize the training data even more precisely without improving generalization to unseen data. However, as the \texttt{Sand} dataset is much larger in size, \texttt{KAN-64} achieves lower test errors exhibiting no overfitting and outperforms all models including \texttt{MLP-256}. As \texttt{KAN-64} training loss is consistently lower in both problems, we can conclude that complex neural architectures such as GNS can benefit from additional expressivity of shallow KANs over MLPs. 

Training time, while not typically a concern for smaller-scale problems, can be significant for GNS models. We report the training times in Table \ref{tab:GNS_training_times}. Notably, \texttt{KAN-64} incurs a substantially higher training cost compared to its MLP counterparts. MLP models rely primarily on simple matrix computation and are highly optimized for modern hardware compared to KANs. Therefore, in general, KAN-based architectures trail in runtime compared to MLP models.

\begin{figure}[ht]
    \centering
    \includegraphics[width=\linewidth]{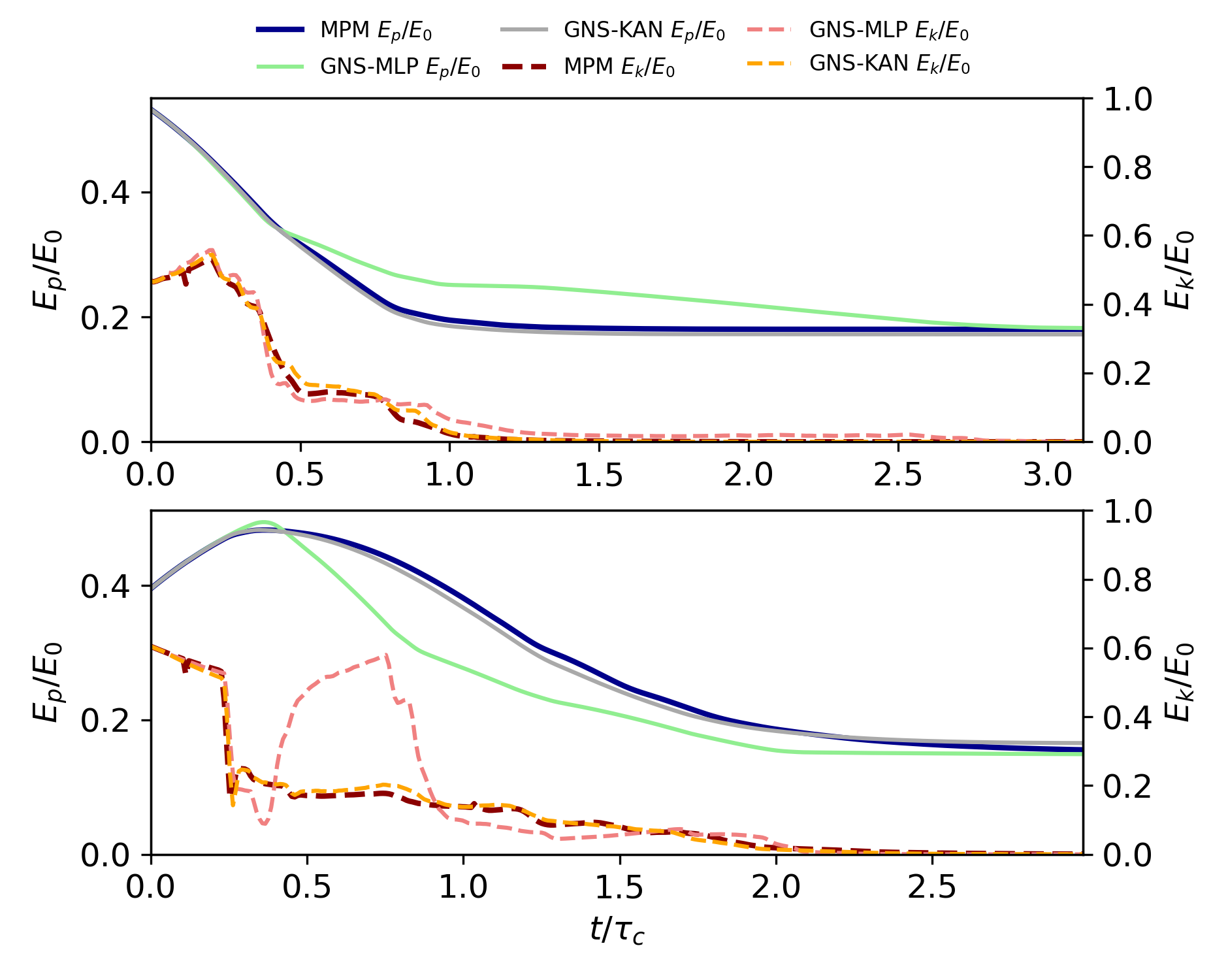}
    \vspace{-1cm}
    \caption{Normalized kinetic and potential energy plots for two test rollouts. The x-axis represents the ratio of elapsed time $t$ to critical time $\tau_c$. GNS represented with KAN outperforms \texttt{MLP-128-128} in predicting physically correct behavior as shown by strong agreement between GNS-KAN and ground truth MPM curves.}
    \label{fig:normalized_energy_subplots}
\end{figure}

A simulated trajectory using both MLP and KAN-based models is shown in \cref{fig:example_gns_mlp_rollout} and \cref{fig:example_gns_kan_rollout} respectively. Two blocks of sand are pushed towards the left and bottom boundaries respectively, and their evolution in time is qualitatively compared against the ground truth MPM data. We note that even though these models are not trained on this trajectory, the predictions strongly agree with ground truth data. 

\begin{figure}[ht]
    \centering
    \includegraphics[width=\linewidth]{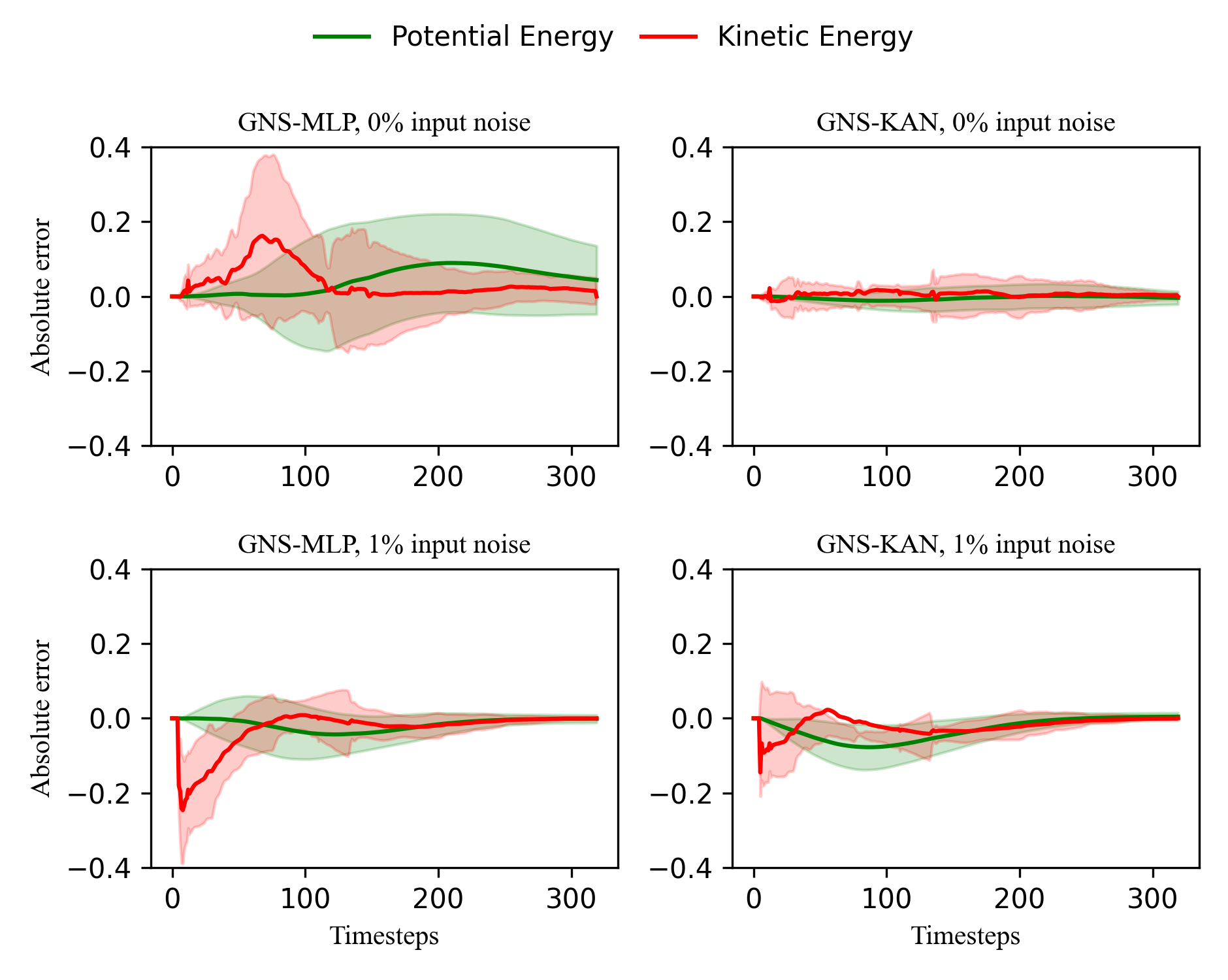}
    \vspace{-1cm}
    \caption{Normalized energy error statistics for MLP and KAN. Top: Standard input. Bottom: Input position sequence perturbed with 1\% noise. Shaded region represents the $99$th percentile of test error.}
    \label{fig:combined_energy_errors}
\end{figure}

Subsequently, we evaluate physical correctness and robustness of model predictions. To that end, we compute normalized kinetic and potential energy of predictions at every time step. Here, we define ``normalized'' as ratio of (kinetic or potential) energy at the current time step to initial potential energy in the system. A total of $25$ unseen trajectories are tested. Energy evolution in two of these $25$ test cases is presented in \cref{fig:normalized_energy_subplots}. The x-axis represents the ratio between elapsed time and critical time $\tau_c$ where the critical time depends on the initial height of the sand column. The flow is fully mobilized when $t/\tau_c = 1$ and comes to a stop at approximately $t/\tau_c = 3$. It can be seen that \texttt{KAN-64} outperforms \texttt{MLP-128-128} models' predictions, i.e. energy of trajectory predicted with \texttt{KAN-64} closely matches the ground truth. Error statistics over all $25$ rollouts are shown in \cref{fig:combined_energy_errors}. To evaluate robustness of the learned model, we add zero-mean Gaussian noise with variance equal to $1\%$ of the initial mean particle velocities to particle positions at the first five time steps that are given as inputs to the GNS models for rollout. It can be seen that \texttt{KAN-64} achieves significantly lower error compared to \texttt{MLP-128-128} with or without noise in the inputs. This study suggests that not only does the GNS represented with KAN learn the particle dynamics more accurately under similar training sample complexity, but it also shows more robustness to noise in input data.

\section{Conclusion}
\label{sec:conclusion}
In this work, we analyze the performance of Kolmogorov–Arnold Networks (KAN) and Multilayer Perceptrons (MLP) in physics-based machine learning tasks. To ensure a comprehensive evaluation, we conduct extensive testing across a diverse set of examples, spanning applications in both operator learning and graph network-based simulators. Our experiments cover problems of varying scales — for instance, we use DeepONet primarily for smaller-scale problems, and leverage GNS to simulate large-scale, particle-based physics over extended time horizons. Our findings provide valuable insights into the strengths and limitations of KAN in both shallow and deep network model configurations.
\begin{itemize}
    \item \textbf{Shallow Neural Network Models: }  KANs consistently outperform MLPs in all but one test scenario when using shallow networks ($1$ layer), demonstrating their superior representation power in these settings. Notably, this performance advantage is observed across both DeepONet and GNS problems, in terms of accuracy as well as robustness to noise in input data, highlighting the versatility of KAN for problems of different scales. Additionally, the reduced size of shallow KAN models leads to a lower computational overhead compared to the deeper networks. Combined with their strong performance, this positions KANs as a practical option for physics applications where small or shallow neural networks are commonly used.  Examples include physics-informed neural networks (PINNs)~\citep{huang2024adaptive, shi2024physics}, kernel methods for linear operator learning~\citep{lin2025orthogonal} and other related approaches. While our findings demonstrate the promise of KAN in this context, we leave a more comprehensive exploration of these methods for future work.
    \item \textbf{Deep Neural Network Models: } In contrast, we do not observe distinct advantages of deep KAN models over MLPs in our testing. While MLP models benefit from increased depth, simply stacking additional KAN layers does not result in comparable performance improvements. This suggests a possible limitation in how KANs capture complex representations when scaled to deeper architectures. Additionally, the computational overhead for KAN models also increases as the number of layers increases, further reducing their appeal in large-scale applications.
    The observed performance stagnation may indicate challenges in optimizing deep KANs, and may suggest that training strategies designed for MLPs may not effectively translate to KAN-based architectures. The nature of the physics-based tasks examined in our study may play a role in these findings. For problems that are well-approximated with lower-dimensional representations, the extra capacity of deep KANs may not provide significant advantages. 
\end{itemize}

Overall, our findings suggest that while KANs offer compelling advantages in shallow network model settings, their effectiveness in deeper architectures remains an open question. The lack of performance gains with increased depth highlights the need for a deeper understanding of the structural properties of KANs and their interactions with model complexity. A deeper study into the optimization of KAN models in physics-related applications could yield valuable insights to help address the problem. Furthermore, exploring hybrid approaches that combine the strengths of both KANs and MLPs could offer promising avenues for enhancing model performance across a broader range of physics-based machine learning tasks. Ultimately, these directions present exciting opportunities for future research in unlocking the full potential of KANs.

\section{Acknowledgment}
This work was partly supported by the National Science Foundation under Grant No. 2321040 and 2339678. Any opinions, findings, and conclusions or recommendations expressed in this material are those of the author(s) and do not necessarily reflect the views of the National Science Foundation.

\section{Data availability}
All the code and data to reproduce results in this paper are available at the Github repositories: \href{https://github.com/geoelements-dev/mlp-kan}{https://github.com/geoelements-dev/mlp-kan} and \\ \href{https://github.com/geoelements/gns/tree/kan-gns}{https://github.com/geoelements/gns/tree/kan-gns}.

\bibliographystyle{elsarticle-harv}
\bibliography{references} 

\end{document}